\documentclass[12pt]{article}

\usepackage{booktabs}       
\usepackage{nicefrac}       
\usepackage{microtype}      
\usepackage{amsfonts,amstext,amsmath,amssymb,amsthm}
\usepackage{smile}
\usepackage{algorithm}
\usepackage{algorithmic}
\usepackage{wrapfig}
\usepackage{lipsum}
\usepackage{ragged2e}
\usepackage{geometry}
\usepackage{setspace}   
\usepackage[symbol]{footmisc}
\renewcommand*{\thefootnote}{\fnsymbol{footnote}}

\usepackage[title,toc]{appendix}
\usepackage{minitoc}
\noptcrule

\usepackage{natbib}
\usepackage[utf8]{inputenc} 
\usepackage[T1]{fontenc}    
\usepackage{hyperref}   
\hypersetup{
	colorlinks=true,       
	linkcolor=blue,        
	citecolor=blue,        
	filecolor=magenta,     
	urlcolor=blue
}
\usepackage{url}            
\usepackage{booktabs}       
\usepackage{amsfonts}       
\usepackage{nicefrac}       
\usepackage{microtype}      
\usepackage[table,xcdraw]{xcolor}
\usepackage{smile}
\usepackage{algorithm}
\usepackage{algorithmic}
\usepackage[]{graphicx,float,latexsym}
\usepackage{amsfonts,amstext,amsmath,amssymb,amsthm}
\usepackage{wrapfig}
\usepackage{lipsum}
\usepackage{gensymb}
\usepackage{quoting,xparse}
\NewDocumentCommand{\bywhom}{m}{
  {\nobreak\hfill\penalty50\hskip1em\null\nobreak
   \hfill\mbox{\normalfont(#1)}%
   \parfillskip=0pt \finalhyphendemerits=0 \par}%
}

\NewDocumentEnvironment{pquotation}{m}
  {\begin{quoting}[
     indentfirst=false,
     leftmargin=\parindent,
     rightmargin=\parindent]\itshape}
  {\bywhom{#1}\end{quoting}}

\usepackage{xcolor}

\newcommand{\newn}{\color{gray}$\quad \star$\color{black}}
\newcommand{\newp}{\color{blue}$\quad +$\color{black}}
\newcommand{\newm}{\color{red}$\quad -$\color{black}}
\newcommand{\newnewn}{\color{gray}$\star$\color{black}}
\newcommand{\newnewp}{\color{blue}$+$\color{black}}
\newcommand{\newnewm}{\color{red}$-$\color{black}}

\newcommand{\newmm}{\color{red}$\boldsymbol{- -}$\color{black}}
\newcommand{\lowernum}[1]{\small{$_{(\text{#1})}$}}

\newcommand{\myt}{\texttt{T}}
\newcommand{\mys}{\texttt{S}}
\def\T{{ \mathrm{\scriptscriptstyle T} }}
\newcommand*{\eg}{\emph{e.g.}{}}
\newcommand*{\ie}{\emph{i.e.}{}}

\newcommand{\cupo}{CUPO}

\def\mytitle{Assessing Electricity Service Unfairness with \\ Transfer Counterfactual Learning}

\newcommand{\independent}{\perp\!\!\!\!\perp} 

\title{\mytitle}
\author{\\ Song Wei$^\dagger$, \ Xiangrui Kong$^\ddagger$, \ Alinson Santos Xavier$^\S$,\\ Shixiang Zhu$^\ddagger$, \ Yao Xie$^\dagger$\footnote{Author correspondence to Yao Xie, e-mail: \texttt{yao.xie@isye.gatech.edu}} , \ Feng Qiu$^\S$\\
\\
  \small{$^\dagger$Georgia Institute of Technology, \quad $^\ddagger$Carnegie Mellon University, \quad  $^\S$Argonne National Laboratory.}
}
\date{\vspace{-20pt}}

\begin{document}

\maketitle

\vfill

\begin{abstract}
Energy justice is a growing area of interest in interdisciplinary energy research. However, identifying systematic biases in the energy sector remains challenging due to confounding variables, intricate heterogeneity in counterfactual effects, and limited data availability. First, this paper demonstrates how one can evaluate counterfactual unfairness in a power system by analyzing the average causal effect of a specific protected attribute. Subsequently, we use subgroup analysis to handle model heterogeneity and introduce a novel method for estimating counterfactual unfairness based on transfer learning, which helps to alleviate the data scarcity in each subgroup. 
In our numerical analysis, we apply our method to a unique large-scale customer-level power outage data set and investigate the counterfactual effect of demographic factors, such as income and age of the population, on power outage durations. Our results indicate that low-income and elderly-populated areas consistently experience longer power outages under both daily and post-disaster operations, and such discrimination is exacerbated under severe conditions. These findings suggest a widespread, systematic issue of injustice in the power service systems and emphasize the necessity for focused interventions in disadvantaged communities.
\end{abstract}


\vfill

\doparttoc 
\faketableofcontents 

\part{} 


\renewcommand*{\thefootnote}{\arabic{footnote}}

\newpage

\section{Introduction}\label{sec:intro}
Energy justice \citep{jenkins2016energy,NCSL2022}---conceptualized as the equitable distribution of energy's benefits and burdens across society---emerges as a focal point in contemporary energy research \citep{mundaca2018successful,roman2019satellite,keady2021energy,heffron2022applying,bhattacharyya2023data,ganz2023socioeconomic,shah2023inequitable,coleman2023energy}. Central to the discourse on energy justice is to detect and rectify systematic biases within electricity service systems, which may stem from policies, infrastructure planning, and resource allocation. These biases exacerbate disparities in energy access and affordability, disproportionately affecting diverse societal groups. As shown by \citet{coleman2020equitable}, various societal groups may exhibit distinct resilience levels to prolonged power outages, with extended outages having potentially devastating impacts on vulnerable populations such as the elderly \citep{nord2006seasonal} and the poor \citep{klinger2014power}.

Recently, there has been increased number of research works \citep{roman2019satellite,ganz2023socioeconomic,shah2023inequitable,coleman2023energy} (even news reports\footnote{\citet{DVORKIN2021}: ``\emph{On 4 August 2020, a tropical storm knocked out power in many parts of New York City ... There are technical reasons that contributed to faster repairs in Manhattan, but in general, the neighborhoods that waited the longest to have their power restored tended to be poorer}.''}) exploring the statistical correlation between socio-demographic factors, such as income and age, and the duration of power restoration during extreme weather events.  
For example, \citet{roman2019satellite} suggests that poorer neighborhoods in Puerto Rico generally experienced longer delays in getting their power back:
\begin{pquotation}{\citet{roman2019satellite}}
{``After Hurricane Maria, a disproportionate share of long-duration power failures ($> 120$ days) occurred in rural municipalities ($41\%$ of rural municipalities vs. $29\%$ of urban municipalities) ... For many urban areas, poor residents, the most vulnerable to increased mortality and morbidity risks from power losses, shouldered the longest outages because they lived in less dense, detached housing where electricity restoration lagged.''}
\end{pquotation}

\begin{figure}[!htp]
\centerline{ 
\includegraphics[width = \textwidth]{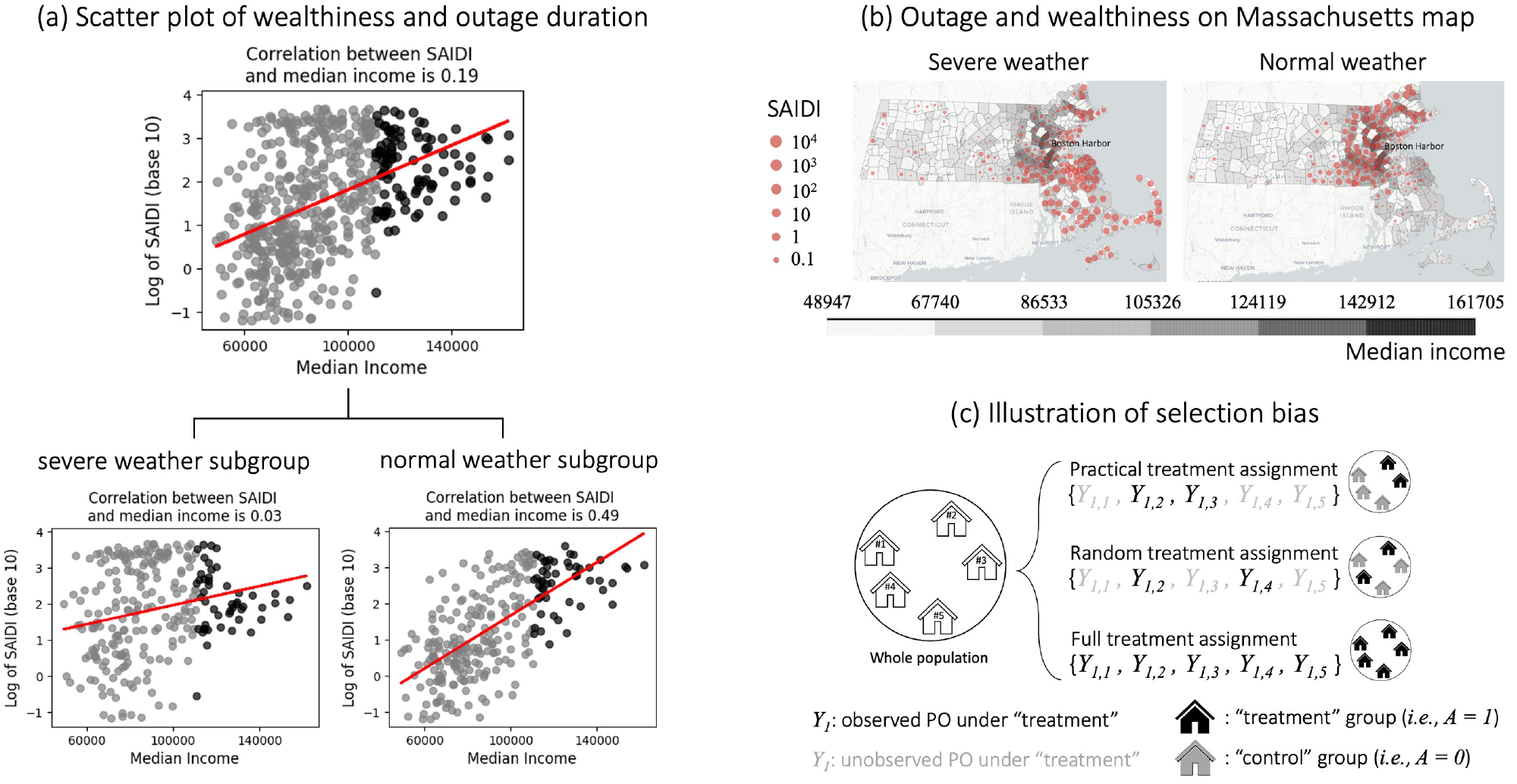} 
}
\vspace{-0.1in}
\caption{\small Evidence and illustration of the selection bias caused by confounders. Sub-figure (a) shows the scatter plot of raw median income---with the lower $80\%$-percentile shaded in gray and the upper $20\%$-percentile in black---and SAIDI, with each panel's trend captured by a red fitted regression line. Although the data visualization on the Massachusetts map in sub-figure (b) indicates lower SAIDI in wealthier areas (especially the Boston Harbor) under severe conditions, the statistical correlation in sub-figure (a) says otherwise.
Sub-figure (c) illustrates the selection bias due to potential confounders in $\boldsymbol{X}$: Commonly in practice (top), the allocation of ``treatment'' is influenced by $\boldsymbol{X}$, making the selected (or observed) ``treatment'' cohort NOT independent of the outcome variable. As a result, this selected cohort fails to reflect the entire population accurately, leading to potentially biased inferences. In our study, as exhibited in sub-figures (a) and (b), we postulate the presence of selection bias, which could account for the discrepancies between the naïve correlation outcomes and the findings of prior studies \citep{roman2019satellite,DVORKIN2021,bhattacharyya2023data,ganz2023socioeconomic,shah2023inequitable,coleman2023energy}, necessitating the counterfactual approach here.}
\label{fig:counterfactual_illus}
\end{figure}

However, three significant gaps persist: 
(1) The reliability of electricity service in normal conditions is vastly under-studied due to a lack of outage data. For example, in Massachusetts, the average outage number per hour during the three consecutive winter storms in March 2018 was $129,012$, accounting for $4.68\%$ of the total customers, with the peak outage number hitting $453,240$. However, during the daily operations in the same month, the average outage number per hour was merely $1,596$, accounting for $0.6\%$ of the total customers and only $1\%$ of that under the storms. 
Such data sparsity makes identifying disparities in standard operations more challenging, which is essential in energy justice literature since it highlights a potential implicit bias in power systems and the need for more systematic targeted support in affected communities. 
(2) Statistical correlations found in existing studies do not necessarily imply causation due to possible confounding biases.
For example, our preliminary analysis 
 illustrated by Figure~\ref{fig:counterfactual_illus} (a) and (b) draws a conclusion that starkly contrasts with previous research \citep{roman2019satellite,DVORKIN2021,bhattacharyya2023data}, \ie, wealthier communities endure longer power outages.
This counter-intuitive correlation indicates the presence of possible exogenous influences from the confounding variables, such as policy choices. 
(3) Model heterogeneity may add another layer of complexity to the assessment of counterfactual unfairness.
Figure~\ref{fig:counterfactual_illus} (a) shows that the impact of a particular policy or infrastructure change might vary widely across different regions, populations, and even weather conditions, making it challenging to generalize findings or draw overarching conclusions.

To tackle these challenges, this paper develops a novel transfer counterfactual learning framework to investigate the unfairness issues in the electricity service systems. 
We first introduce a new notion under Rubin's Potential Outcome (PO) framework \citep{rubin1974estimating}, termed \emph{Counterfactual Unfairness under Potential Outcome} (\cupo). 
We argue that the \cupo \ is better suited for this unfairness assessment task, because 
(1) The estimation of \cupo \ can statistically remove the potential confounding biases, as illustrated by Figure~\ref{fig:counterfactual_illus} (c), 
(2) Identifying \cupo \ can suggest the system is counterfactually unfair \citep{kusner2017counterfactual} to the protected attributes 
under Pearl's Structural Causal Model (SCM) framework \citep{pearl2009causality}, and 
(3) The assessment of \cupo \ can be easily implemented by the average causal effect estimation, which is more reliable and less challenging than inferring the entire causal graph under the SCM framework.
To address the issues of model misspecification caused by heterogeneous counterfactual effects across subgroups and the challenge of data scarcity,
we develop \emph{$\ell_1$ regularized Transfer Counterfactual Learning ($\ell_1$-TCL)}, where we incorporate a recent $\ell_1$ regularized transfer learning technique \citep{bastani2021predicting,cai2022transfer,lin2023source} within the inverse probability weighting (IPW) \citep{horvitz1952generalization} estimation to rectify nuisance model biases \citep{wei2023transfer} and enhance the causal effect estimation accuracy. 
Moreover, we provide a non-asymptotic theoretical guarantee on the estimation error with the help of Lasso in the high-dimensional setting \citep{candes2007dantzig,bickel2009simultaneous} and an empirical scheme on regularization hyperparameter selection by assessing the covariate balance with Maximum Mean Discrepancy \citep{gretton2012kernel}.
Lastly, we apply the $\ell_1$-TCL to a unique large-scale city-level outage dataset, covering $351$ cities in Massachusetts (MA) from March 1st to March 31, 2018\footnote{In March 2018, three successive Nor'easters, \ie, storms along the East Coast of North America, struck the Northeastern United States, resulting in economic damages exceeding $\$729,000$ \citep{ibm2018,wikipedia2018}. 
In particular, the third winter storm, unofficially named Skylar, unleashed severe blizzard conditions across MA, bringing wind gusts around $55$-$65$ miles per hour and over $28$ inches snowfall totals. These adverse events led to large-scale power outages that affected over $230,000$ customers in MA \citep{ABC2018}.}. Our analysis indicates disadvantaged communities, \eg, the economically-challenged and elderly-populated, endure extended outages, irrespective of the weather conditions, suggesting systemic biases rooted in the power service distribution systems and underscoring the necessity for targeted interventions in those vulnerable sectors of the society.

To summarize, the main contribution of this work is three-fold:
\begin{itemize}
    \item[1.] \textbf{What-is}: We demonstrate that the counterfactual unfairness can be assessed by estimating the average causal effect under the Potential Outcome framework.
    \item[2.] \textbf{How-to}: We develop a novel transfer counterfactual learning framework that addresses the confounding bias, model heterogeneity, and data scarcity issues.
    We also provide a non-asymptotic theoretical guarantee on the estimation error and an empirical regularization hyperparameter selection scheme based on covariate balances.
    \item[3.] \textbf{Impact}: 
    We perform the transfer counterfactual analysis on a unique large-scale dataset. The analysis indicates a clear pattern that communities with lower income or a higher elderly population percentage experience more frequent and longer electricity outages, in normal conditions and during extreme weather.
\end{itemize}

\section{Related Work}

\paragraph{Counterfactual analysis for fairness assessment.} 
Fairness with respect to (w.r.t.) the data generation process (DGP), referred to as the \emph{DGP fairness}, is typically the first step towards developing fair strategy (\eg, a predictive model) and making fair impacts in the long term \citep{tang2023and}.
However, identifying systematic biases in the DGP is challenging due to confounding attributes, leading to the recent development of the counterfactual analysis. 
Indeed, counterfactual modeling is a rather straightforward choice for DGP fairness assessment since ``one of the strongest motivations behind the usage of a causal notion is the insight into the DGP behind the outcome $Y$ in the current reality'' \citep{tang2023and}. As illustrated in Figure~\ref{fig:counterfactual_illus}, the DGP fairness assessment in our real example needs the help of counterfactual analysis. Unfortunately, such kind of analysis is largely missing in the Energy Justice literature.

\paragraph{Transfer learning for counterfactual analysis under PO framework.} 
While there has been increased interest in applying data-integrative Transfer Learning (TL) techniques to the counterfactual analysis under the PO framework (\ie, \emph{causal inference}) in the presence of heterogeneous covariate spaces \citep{yang2020combining,wu2022transfer,hatt2022combining,bica2022transfer}, these methods typically fail to handle the same covariate space setting, known as the inductive multi-task transfer learning \citep{pan2010survey}. This limitation arises from their algorithm designs, which mostly rely on domain-specific covariate spaces. 
To the best of our knowledge, the first and only work studying data-integrative TL for causal effect estimation under the inductive multi-task setting is \citet{kunzel2018transfer}. They proposed to transfer knowledge by using neural network (NN) weights estimated from the source domain as the warm start of the subsequent target domain NN training. Despite its improved empirical performance, the theoretically grounded approach is still largely missing, and such an NN-based TL approach fails to generalize to parametric methods such as (generalized) linear models. 

A notable empirical challenge in causal inference is the calibration of the hyperparameter, as \citet{machlanski2023hyperparameter} recently pointed out that proper hyperparameter tuning can significantly narrow the performance gap among various causal estimators. However, in contrast to model selection in conventional supervised learning, causal inference lacks a direct counterpart to cross-validation due to the inability to assess causal effect estimation accuracy, given the unobservable nature of counterfactual outcomes. Presently, the absence of a theoretically robust criterion has led to reliance on empirical analyses, often utilizing the performance of auxiliary nuisance models (\eg, the propensity score model) as selection criterion \citep{athey2016recursive,curth2023search,machlanski2023hyperparameter,mahajan2024empirical}. In response to this gap, we introduce an additional approach by examining covariate distribution balance and evaluating the efficacy of these criteria through both numerical simulations and our real-world case study.

\section{Counterfactual Unfairness Assessment under the Potential Outcome Framework}
We start with clarifying the fairness notion used for unfairness assessment in our real application: the group-level, static, counterfactual unfairness under Rubin's Potential Outcome framework, \ie, \cupo. We formally define the counterfactual fairness and unfairness notions under the PO framework, which can be estimated via the average causal effect. Importantly, we will elaborate on the connections between the counterfactual (un)fairness notions under both PO and SCM frameworks, and explain why our \cupo \ is better suited for our motivating real application.

\subsection{Fairness spectra}
Following \citet{tang2023and}, we differentiate fairness notions w.r.t. data generating process, predicted outcome, and induced impact. This work focuses on DGP fairness, which inquires ``the existence of discrimination in the data that results from the imperfection of previous committee decisions ... without considering downstream tasks'' \citep{tang2023and}. Auditing DGP fairness usually precedes training ``fair'' prediction algorithms, which usually (implicitly) assumes the data itself is ``clean'' (according to the bias definition of interest). As one can see, the DGP fairness reflects fairness in the existing policy that defines the DGP, making it the suitable fairness notion in our unfairness assessment application.

Additionally, we target fairness issues under a \emph{static} setting, \ie, neither time-dependent nor long-term effect \citep{d2020fairness} is of our consideration. This is a mild and plausible assumption since we only focus on the outage data spanning one month in our motivating application. Meanwhile, we do acknowledge that fairness w.r.t. induced impact is of interest in the follow-up study, which could involve long-term effects. For example, one may wonder whether the prolonged power outage in the low-income area (if there were any) will lead to persistent economic challenges in the long term \citep{bohr2020energy}. ``Admittedly, the detection of the existence of discrimination in the data does not easily translate into possible ways to perform correction'' \citep{tang2023and}, nevertheless, DGP fairness assessment is the first step towards fairness w.r.t. both predicted outcome and induced impact.

Lastly, we focus on \emph{group-level fairness} instead of individual-level fairness, since unfairness on a group level is enough to justify discrimination in the existing policy. We will elucidate the reason behind this choice later in Section~\ref{rmk:individual-or-group-level}.

\subsection{Causal notions for fairness}
We study energy justice by asking ``what if'' questions, such as ``What would the power outage duration have been if this city were a wealthy one?'' We answer those questions by studying the causal effect from protected attribute (\eg, wealthiness, elderly percentage, etc., treated as the ``treatment'' variable) to the outcome variable (\ie, the power outage measured by SAIDI). 
In this subsection, we formally introduce the counterfactual fairness and unfairness notions under the Potential Outcome framework. Importantly, we will shed light on how our \cupo \ can suggest a disparity in terms of the well-known \emph{Counterfactual Fairness} due to \citet{kusner2017counterfactual}, which will be referred to as the \emph{SCM-counterfactual fairness} for presentation clarity. 

\subsubsection{Counterfactual (un)fairness under the Potential Outcome framework}
Consider the tuple $(\boldsymbol{X}, {A}, {Y})$, where random variable (r.v.) ${A}\in \{0,1\}$ represents protected attribute (for example, ${A}= 1$ if the median household income is greater than a pre-defined threshold and $0$ otherwise), random vector $\boldsymbol{X} \in \cX \subset \RR^d$ denotes additional (observable) attributes (such as weather condition), and r.v. ${Y}$ is the outcome, \ie, the power outage duration measurement. Under Rubin's PO framework, \[{Y} = {Y}_{{1}} {A}+ (1-{A}) {Y}_{{0}},\] where \({Y}\) is referred to as the \emph{observed outcome}, whereas ${Y}_{{0}}$ and ${Y}_{{1}}$ are called \emph{potential outcomes}---they are the values of the outcome that would be seen if the subject were to receive ``control'' (\ie, ${A} = 0$) or ``treatment'' (\ie, ${A} = 1$). The notion of PO (which is similar to the \( \textit{do}(\cdot) \) operator in Pearl's SCM framework) is important as it helps to deal with confounders in \( \boldsymbol{X} \), allowing for the isolation of the direct effect of \( {A} \) on \( {Y} \) without the influence of confounding factors.

One commonly considered causal estimand is the average causal effect (ACE), formally defined as follows:
\[\text{\rm Average Causal Effect:} \ \tau = \EE[{Y}_{{1}}] - \EE[{Y}_{{0}}].\]
In an observational study, attributes in $\boldsymbol{X}$ that determine (or cause) the treatment (\ie, protected attribute) may also be correlated, or ``confounded'', with the outcome. Thus, selecting subjects with ${A} = 1$ (or $0$) and averaging the selected outcomes will yield a biased estimate of $\EE[{Y}_{{1}}]$ (or $\EE[{Y}_{{0}}]$). Although such statistical independence typically does not hold in practice, a common practice in causal inference literature to handle such a problem is to assume there are ``no unmeasured confounders'':
\begin{equation}\label{assumption:ignorability}
    ({Y}_{{0}}, {Y}_{{1}}) \independent {A} \mid \boldsymbol{X}.
\end{equation}
This assumption is known as the \emph{Ignorability Assumption} \citep{robins2000marginal}, which ensures the ACE is identifiable from an observational study. We shall continue the rest of our study under this assumption. Importantly, based on the notion of ACE, we introduce the DGP fairness notion under the PO framework:
\begin{definition}[Group-level counterfactual fairness under PO] Under the Ignorability Assumption \eqref{assumption:ignorability}, given the tuple $(\boldsymbol{X}, {A}, {Y})$, we say the data generating process yielding outcome ${Y}$ is fair w.r.t. the protected feature \({A}\) in terms of group-level counterfactual fairness under the Potential Outcome framework if the average causal effect is zero, \ie,
\begin{equation}\label{eq:group-fairness_PO}
    \tau = \EE[{Y}_{{1}}] - \EE[{Y}_{{0}}] = 0.
\end{equation}
\end{definition}
We claim existence \cupo \ if the estimated ACE does not satisfy \eqref{eq:group-fairness_PO}; Due to the uncertainty in real applications, we quantify the uncertainty via, \eg, Bootstrap, to assess \cupo \ based on whether the confidence interval contains zero.

\subsubsection{\textit{Connection}: The connection between the potential outcome and \( \textit{do}(\cdot) \) operator translates into the bridge connecting the counterfactual fairness notions under the PO and SCM frameworks}
As a recap, we briefly introduce Pearl's SCM framework, where the DGP is modeled via \((\boldsymbol{U}, \boldsymbol{V}, \boldsymbol{F})\): \(\boldsymbol{V} = (\boldsymbol{X}, {A}, {Y})\) is the set of observable variables, \(\boldsymbol{U}\) is a set of unobserved variables which are not caused by any variable in \(\boldsymbol{V}\), and \(\boldsymbol{F}\) is a set of structural equations characterizing how observable variables are generated; We refer readers to \citet{kusner2017counterfactual} for complete details of SCM-counterfactual fairness. Notably, \citet{tang2023and} extends the SCM-counterfactual fairness to handle the DGP fairness:
\begin{definition}[SCM-counterfactual fairness for the DGP \citep{tang2023and}]
Given a causal model \((\boldsymbol{U}, \boldsymbol{V}, \boldsymbol{F})\) that describes the data generating process of the current reality, where we assume there are no unobserved confounders \eqref{assumption:ignorability}, \ie, \(\boldsymbol{U} = \emptyset\), and \(\boldsymbol{V} = (\boldsymbol{X}, {A}, {Y})\), we say that the outcome ${Y}$ is fair w.r.t. the protected feature \({A}\) in terms of counterfactual fairness under the SCM framework, if for any \( \boldsymbol{X} = \boldsymbol{x} \) and \( {A} = a \), the following condition holds:
\begin{equation}\label{eq:CF}
    \mathbb{P}({Y}_{\textit{do}({A}) = a} = y | \boldsymbol{X} = \boldsymbol{x}, {A} = a) - \mathbb{P}({Y}_{\textit{do}({A}) = a'} = y | \boldsymbol{X} = \boldsymbol{x}, {A} = a) = 0,
\end{equation}
for all \( y \) and any values \( a' \) that \( A \) could possibly take. 
\end{definition}

In the above definition, SCM-counterfactual fairness utilizes the \( \textit{do}(\cdot) \) operator to denote the hypothetical intervention of setting \( {A} \) to \( a' \), which conceptually simulates the changes to the protected attribute while controlling for other variables in \( \boldsymbol{X} \). 
This is the bridge connecting Pearl's SCM framework with Rubin's PO framework, and the key idea under both frameworks is to conduct the counterfactual analysis by deriving the connections between the (unobserved) interventional and observed distributions \citep{malinsky2019potential}. 
They merely use different notions (\ie, potential outcomes and \textit{do}$(\cdot)$ operator, respectively) to formalize the counterfactuals.
As Pearl himself mentioned: ``Queries about causal
effects (written \(\mathbb{P}({Y}_{\textit{do}({A}) = a} = {y})\) in the structural analysis) are phrased as queries about the marginal distribution of the counterfactual variable of interest, written \(\mathbb{P}({Y}_{a} = {y})\)'' \citep{pearl2010causal}.

\subsubsection{\textit{Difference}: ``Group or individual level'' is NOT the key difference between counterfactual fairness under PO and SCM frameworks}\label{rmk:individual-or-group-level}
As a brief recap of the counterfactual fairness development, the famous notion of \emph{Counterfactual Fairness} was first proposed by \citet{kusner2017counterfactual} under Pearl's SCM framework \citep{pearl2009causality} (\ie, the SCM-counterfactual fairness) to assess fairness in the predictive models, and it has been recently extended to handle the fairness in the DGP \citep{tang2023and}. In the meantime,  \citet{khademi2019fairness} proposed to assess the counterfactual fairness in the DGP by estimating the average causal effect under Rubin's PO framework, however, they did not adequately address its connection to the SCM-counterfactual fairness. 
\citet{tang2023and} mentioned that the SCM-counterfactual fairness \citep{khademi2019fairness} is on group-level whereas the commonly known SCM-counterfactual fairness \citep{kusner2017counterfactual} is on individual-level, but one can easily extend the counterfactual fairness under PO to the individual-level using the notion of individual treatment effect:
\begin{definition}[Individual-level counterfactual fairness under PO]
Under the Ignorability Assumption \eqref{assumption:ignorability}, given the tuple $(\boldsymbol{X}, {A}, {Y})$, we say the data generating process yielding outcome ${Y}$ is fair w.r.t. the protected feature \({A}\) in terms of individual-level counterfactual fairness under the Potential Outcome framework, if the individual treatment effect is zero, \ie, for any \( \boldsymbol{X} = \boldsymbol{x} \),
\begin{equation}\label{eq:individual-fairness_PO}
    \tau_{\boldsymbol{x}} = \EE[{Y}_{{1}}|\boldsymbol{X} = \boldsymbol{x}] - \EE[{Y}_{{0}}|\boldsymbol{X} = \boldsymbol{x}] = 0.
\end{equation}
\end{definition}
Thus, ``group- or individual-level'' is not the key difference between the counterfactual fairness notions under the PO and SCM framework. Indeed, both frameworks are ``logically equivalent'', and they are ``two different languages for the same thing'' \citep{Pearl2012}. One key difference between both frameworks is the granularity of the counterfactual effects: While SCM-counterfactual fairness \eqref{eq:CF} tries to infer the whole counterfactual conditional probabilistic distribution, the individual-level counterfactual fairness under PO \eqref{eq:individual-fairness_PO} only characterizes fairness in terms of the first-order moment of that distribution. 
As a result, the counterfactual fairness notions under PO, \ie, \eqref{eq:group-fairness_PO} and \eqref{eq:individual-fairness_PO} are weaker fairness notions compared to the SCM-counterfactual fairness \eqref{eq:CF}, but our \cupo \ is a stronger unfairness notion which can imply the existence of disparity in terms of the SCM-counterfactual fairness notion. To be precise, we have:
\begin{proposition}\label{prop:relashionship}
    SCM-counterfactual fairness \eqref{eq:CF} implies individual-level DGP fairness under the PO framework \eqref{eq:individual-fairness_PO}, which further implies group-level DGP fairness under the PO framework \eqref{eq:group-fairness_PO}.
\end{proposition}

``Depending on the task at hand, one framework might be more appropriate to use than the other'' \citep{makhlouf2020survey}. The difference in the granularity of the counterfactual analysis gives an edge to our \cupo \ notion when the task is to detect the counterfactual unfairness: (1) Proposition~\ref{prop:relashionship} implies that assessment of \cupo \ could suggest SCM-counterfactual unfairness, and (2) inferring (sub)group-level DGP fairness via estimating (conditional) causal effect is much easier and reliable than recovering the entire causal graph under the SCM framework, especially under the limited data setting. As a result, we argue that assessing \cupo \ is better suited for our unfairness detection task in the power service systems.

\section{Transfer Counterfactual Learning}\label{sec:background}
In this section, we will focus on the ``how-to'' part in assessing our \cupo. We will give a brief introduction of the formal set-up in Rubin's PO framework and introduce how we adapt a recent transfer learning technique to causal effect estimation.

\subsection{Inverse probability weighting to mitigate the selection bias}\label{sec:selection_bias}
``When observations in social research are selected so that they are not independent of the outcome variables in a study, sample selection leads to biased inferences about social processes'' \citep{winship1992models}. To help understand the selection bias, we present a graphical illustration in Figure~\ref{fig:counterfactual_illus}(c). Unlike experimental studies where we can manually force every subject to receive treatment or randomly assign treatments, it is usually impossible to manipulate the protected attribute ${A}$, such as household income and senior population percentage in our observational study. As a result, the outcomes in the selected (or observed) ``treatment'' cohort may be influenced by other attributes in $\boldsymbol{X}$ instead of the ``treatment'' ${A}$ itself. 

One way to handle the selection bias is to re-weight each sample in the selected cohort such that the re-weighted sample is ``representative'' of the whole population, and one popular method is the inverse probability weighting \citep{horvitz1952generalization}. To be precise, \citet{rosenbaum1983central} showed that the propensity score $e(\boldsymbol{X}) = \PP({A} = 1|\boldsymbol{X})$, \ie, the probability of receiving ``treatment'' given covariates, satisfies the following:
\[({Y}_{{0}}, {Y}_{{1}}) \independent {A} \mid e(\boldsymbol{X}).\]
This leads to the following unbiased estimate of $\EE[{Y}_{{1}}]$:
\begin{equation}\label{eq:IPW_unbiased}
        \EE\left[\frac{{A} Y}{e(\boldsymbol{X})}\right] = \EE\left\{\EE\left[\frac{I({A} = 1) Y_{{1}}}{e(\boldsymbol{X})} \ \bigg| \ Y_{{1}}, \boldsymbol{X}\right]\right\}  = \EE\left\{\frac{Y_{{1}}}{e(\boldsymbol{X})} \EE\left[I({A} = 1) | Y_{{1}}, \boldsymbol{X}\right]\right\} =\EE[Y_{{1}}].
\end{equation}
Similarly, we have \(\EE\left[\frac{{A} Y}{1-e(\boldsymbol{X})}\right] =\EE[Y_{{0}}]\). Then, the famous inverse probability weighting (IPW) estimator of the ACE $\tau$ can be obtained by replacing the expectation with sample average in $\EE\left[\frac{{A} Y}{e(\boldsymbol{X})}\right] - \EE\left[\frac{{A} Y}{1-e(\boldsymbol{X})}\right]$; We will introduce the IPW estimator in detail later. Following standard terminology in causal inference, we will call \(\boldsymbol{X}\) (pre-treatment) covariates and \({A}\) treatment (indicator) in the following study.

\subsection{Subgroup analysis to handle model heterogeneity}
One challenge faced in this work is the potential heterogeneity, \ie, the treatment assignment rule and/or the true causal effect could vary among different subgroups within a population due to diverse individual characteristics and responses. The latter case is referred to as \emph{Heterogeneous Causal Effect (HCE)} problem, under which the ACE for the whole population is less meaningful. One common method to mitigate the impact of HCE is through studying the conditional average causal/treatment effect (CATE) instead of ACE within each subgroup, \ie,
\[\text{\rm Conditional ACE: } \tau_{S} = \EE[ {Y}_{{1}} | \boldsymbol{X} \in \cS ] - \EE[ {Y}_{{0}} | \boldsymbol{X} \in \cS ].\]
In our study, the subgroup is obtained as ``covariates lying in a target subset of the covariate space'', \ie, $\boldsymbol{X} \in \cS \subset \cX$. Even though there exists a data-driven approach to partition the population into subgroups, such as \citet{athey2016recursive}, we focus on a more interpretable approach that is based on the label of a binary (or categorical) covariate; Specifically, in this work, we partition the data into normal and extreme weather subgroups based on the whether condition (such as wind and rainfall). Without loss of generality, we consider that the first element in the pre-treatment covariate vector is binary, \ie, $\boldsymbol{X} = (X_1, X_2,\dots), X_1 \in \{0,1\}$. This yields a natural partition that 
\begin{equation}\label{eq:partition}
    \cX = \cS_{\mys} \cup \cS_{\myt}, \ \text{\rm where } \cS_{\mys} = \{X_1 = 0\} \ \text{\rm and } \cS_{\myt} = \{X_1 = 1\}.
\end{equation}
In the HCE problem, we typically have a heterogeneous causal effect, \ie, \(\tau_{\cS_{\mys}} \not= \tau_{\cS_{\myt}}.\)
Thus, we are interested in estimating CATE, which is typically done by estimating ACE using the corresponding subgroup of the observations whose covariates lie in $\cS_{\mys}$ or $\cS_{\myt}$.

\subsection{Transfer learning to handle data scarcity}
The subgroup analysis above typically suffers from insufficient sample size. To handle this problem, we consider transfer learning techniques by leveraging knowledge obtained from one subgroup (called \emph{source domain}) to help the estimation in the other subgroup (called \emph{target domain}).
Existing approaches for HCE estimation largely focus on model-ensemble, such as meta-learning \citep{curth2021nonparametric}, and heterogeneous transfer learning (\ie, TL under the heterogeneous covariate space setting) \citep{bica2022transfer}, but is not applicable to our problem set-up. Here, we adapt a recent $\ell_1$ regularized TL approach due to \citet{bastani2021predicting,wei2023transfer} to estimate HCE.

\subsubsection{Preliminaries}
We call $\cS_{\myt}, \cS_{\mys}$ \eqref{eq:partition} the target domain and source domain, respectively. 
Due to the potential heterogeneity, we denote $(\boldsymbol{X}_{\myt}, {A}_{\myt}, Y_{\myt})$ and $(\boldsymbol{X}_{\mys}, {A}_{\mys}, Y_{\mys})$ as the random vector under the corresponding domains, and our observations are 
\begin{equation*}
\begin{split}
    \text{Target Domain: } \cD_{i,\myt} &= (\boldsymbol{x}_{i, \myt}, {a}_{i, \myt}, y_{i, \myt}), \ i = 1,\dots, n_{\myt}, \text{ where } \ \boldsymbol{x}_{i, \myt} \in \cS_{\myt},\\
    \text{Source Domain: } \cD_{i,\mys} &= (\boldsymbol{x}_{i, \mys}, {a}_{i, \mys}, y_{i, \mys}), \ i = 1,\dots, n_{\mys}, \text{ where } \ \boldsymbol{x}_{i, \mys} \in \cS_{\mys}.   
\end{split}
\end{equation*}
To mitigate the selection bias, we apply the IPW estimator to estimate the ACE, which requires estimating the propensity score. We consider a simple yet popular generalized linear form for propensity score models:
\begin{equation*}
    \begin{split}
        \PP({A}_{\myt} = 1 | \boldsymbol{X}_{\myt}) = g({\boldsymbol{X}_{\myt}}^\T \beta_{\myt}), \quad \PP({A}_{\mys} = 1 | \boldsymbol{X}_{\mys}) = g({\boldsymbol{X}_{\mys}}^\T \beta_{\mys}),
    \end{split}
\end{equation*}
where superscript $^\T$ denotes vector or matrix transpose, and $g(\cdot)$, known as the (inverse) link function, can be either linear, \ie, $g(x) = x, \ x\in [0,1]$, or nonlinear, such as sigmoid link function $g(x) = 1/(1+e^{-x}), \ x \in \RR,$ and exponential link function $g(x) = 1 - e^{-x}, \ x \in [0, \infty)$. 
The IPW estimator for CATE (or ACE in the subgroup) is defined as:
\begin{equation}\label{eq:ipw}
    \hat \tau_{\myt} = \frac{1}{n_\myt} \sum_{i = 1}^{n_{\myt}} \frac{{a}_{i,\myt} y_{i,\myt}}{g(\boldsymbol{x}_{i,\myt}^\T \hat \beta_{\myt})} - \frac{(1-{a}_{i,\myt}) y_{i,\myt}}{1-g(\boldsymbol{x}_{i,\myt}^\T \hat \beta_{\myt})}, \quad 
    \hat \tau_{\mys} = \frac{1}{n_\mys} \sum_{i = 1}^{n_{\mys}} \frac{{a}_{i,\mys} y_{i,\mys}}{g(\boldsymbol{x}_{i,\mys}^\T \hat \beta_{\mys})} - \frac{(1-{a}_{i,\mys}) y_{i,\mys}}{1-g(\boldsymbol{x}_{i,\mys}^\T \hat \beta_{\mys})},
\end{equation}
where $\hat \beta_{\myt}, \hat \beta_{\mys}$ are the estimated nuisance parameters in the target and source domains, respectively.

In the above generalized linear model (GLM) formulation, we consider the same link function $g(\cdot)$ but different model parameters $\beta_{\myt} \neq \beta_{\mys}$, which accounts for related yet different domains; That is to say, our formulation can account for not only HCE but also heterogeneous treatment assignment rules. 

\begin{remark}[Further clarification on the TL set-up]
    Our set-up falls into the category of ``inductive multi-task transfer learning'' according to \citet{pan2010survey}. To be precise, we consider homogeneous feature space but (potentially) heterogeneous feature distribution, treatment assignment rule, and causal effect across the source and target domains.
\end{remark}

\subsubsection{Proposed method} 
Since most transfer learning problems fall in the category of supervised learning, it is natural to consider knowledge transfer for the propensity score estimation stage in the IPW estimation. In our setting, we consider improving the estimation accuracy of $\beta_{\myt}$ with the knowledge gained on $\beta_{\mys}$, which helps resolve the data insufficiency issue due to partition. The key assumption for successful, \ie, theoretically guaranteed, knowledge transfer is the sparsity of the nuisance parameter difference \citep{bastani2021predicting,wei2023transfer}; Formally, the nuisance parameter difference $\Delta_\beta$, defined as:
\begin{equation}\label{eq:beta_diff}
    \Delta_\beta = \beta_\myt - \beta_\mys,
\end{equation}
is assumed to be $s$-sparse, \ie, for $0 \leq s \leq d$, the vector $\ell_0$ norm of $\Delta_\beta$ satisfies: 
\begin{equation}\label{assumption}
    \norm{\Delta_\beta}_0 \leq s.
\end{equation}
This assumption states that treatment assignment mechanisms are very similar across both domains. Please see Figure~\ref{fig:realExpMADelta} for numerical evidence supporting this assumption in our real data example.

To estimate $\beta_\myt$, we first leverage source domain data to estimate $\beta_\mys$, which serves as a rough estimator of $\beta_\myt$ due to Assumption~\eqref{assumption}. Next, we correct the bias of the rough estimator by using $\ell_1$ regularization to learn the difference $\Delta_\beta$ from target domain data. The idea is that we can accurately estimate $\beta_\mys$ using abundant source domain data and faithfully capture the sparse $\Delta_\beta$ from limited target domain data with the help of $\ell_1$ regularization. To be precise:
\begin{itemize}
    \item[] Rough estimation: \(\hat \beta_\myt^{\ \rm rough} = \arg \min_b \frac{1}{n_\mys} \sum_{i=1}^{n_\mys} -{a}_{i, \mys} \boldsymbol{x}_{i, \mys}^{\T} b+G\left(\boldsymbol{x}_{i, \mys}^{\T} b\right),\)
    \item[] Bias correction: \(\hat \beta_\myt^{\ \ell_1\rm{TCL}} = \arg \min_b \frac{1}{n_\myt} \sum_{i=1}^{n_\myt} -{a}_{i,\myt} \boldsymbol{x}_{i,\myt}^{\T} b+G\left(\boldsymbol{x}_{i,\myt}^{\T} b\right) + \lambda \norm{b - \hat \beta_\myt^{\ \rm rough}}_1,\)
\end{itemize}
where $\lambda > 0$ is a tunable regularization strength hyperparameter, \(\norm{\cdot}_1\) is the vector $\ell_1$ norm, and function $G$ satisfies $G' = g$; Please refer to \citet{nelder1972generalized} for a detailed definition and estimation of the generalized linear model.
Lastly, the proposed $\ell_1$-TCL estimation of $\tau_{\myt}$ can be done by plugging $\hat \beta_\myt = \hat \beta_\myt^{\ \ell_1\rm{TCL}}$ into \eqref{eq:ipw}, \ie,
\begin{equation}\label{eq:TLIPW}
    \begin{split}
        \hat \tau_{\myt}^{\ \ell_1\rm{TCL}} & = \frac{1}{n_\myt} \sum_{i = 1}^{n_{\myt}} \frac{{a}_{i,\myt} y_{i,\myt}}{g(\boldsymbol{x}_{i,\myt}^\T \hat \beta_{\myt}^{\rm TL})} - \frac{(1-{a}_{i,\myt}) y_{i,\myt}}{1-g(\boldsymbol{x}_{i,\myt}^\T \hat \beta_{\myt}^{\rm TL})}. 
    \end{split}
\end{equation}

\begin{remark}[Selection criterion for hyperparameter $\lambda$]\label{rmk:selection_criteria}
   As mentioned earlier, selecting the $\ell_1$ regularization hyperparameter $\lambda$ is crucial \citep{machlanski2023hyperparameter}.
   In this work, in addition to nuisance model prediction metrics studied in literature \citep{athey2016recursive,curth2023search,machlanski2023hyperparameter,mahajan2024empirical}, we introduce another approach by examining covariate distribution balance after the inverse propensity score weighting with the help of the Maximum Mean Discrepancy (MMD) \citep{gretton2012kernel}. Later in Section~\ref{appendix:criteria}, we will introduce these selection criteria and present numerical simulation results on the comparison among these criteria.
\end{remark}

\begin{remark}[Theoretical guarantee]
    As shown by \citet{bastani2021predicting,wei2023transfer}, when the propensity score model is correctly specified, and the difference is \( s \)-sparse \eqref{assumption}, in the large sample limit \( n_{\myt}, n_{\mys} \rightarrow \infty \), consider the following regime:
\begin{equation}\label{eq:regime}
    n_{\myt}  \gg  s^2 \log d, \quad n_{\mys}  \gg  n_{\myt} d^2.
\end{equation}
By taking \( \lambda = \cO\left(\sqrt{\log d} \left(\frac{1}{\sqrt{n_{\myt}}} + \frac{d}{\sqrt{n_{\mys}}}\right)\right), \)
we can show that, with probability at least \( 1- 1/n_{\myt} \), the absolute estimation error can be upper bounded as follows:
\begin{equation*}
    \begin{split}
        \left| \hat \tau_{\myt}^{\ \ell_1\rm{TCL}} - \tau \right| \ = \cO \bigg( \underbrace{s d \sqrt{\frac{\log d}{n_{\mys}}}}_{\text{\rm\emph{rough estimation error}}} + \underbrace{s \sqrt{\frac{\log d}{n_{\myt}}}}_{\text{\rm\emph{bias correction error}}} \bigg),
    \end{split}
\end{equation*}
where we slightly abuse notation \(\tau\) to denote the true ACE in the target domain.
The formal statement of our theory (including proper definitions of the asymptotic notations) and its complete proof can be found in Appendix~\ref{appendix:TLIPW} for completeness purpose. 
The proof approach mirrors the two-stage estimation process, specifically the recovery error of nuisance parameters followed by its substitution into the plug-in estimators to derive the above error bound. 
\end{remark}

\section{Empirical Study on Hyperparameter Selection}\label{appendix:criteria}

One practical issue is the selection of the $\ell_1$ regularization hyperparameter $\lambda$. In fact, this is a rather important topic as proper hyperparameter tuning can sometimes close the performance gap among different causal estimators \citep{machlanski2023hyperparameter}. The challenge is that a ``golden standard'', such as prediction accuracy in classic supervised learning, is largely missing since we cannot evaluate the causal effect estimation accuracy from data due to unobserved counterfactual outcomes. Unfortunately, there is no theoretically grounded solution beyond the recent empirical analysis of some nuisance model prediction performance metrics as the selection criteria \citep{athey2016recursive,curth2023search,machlanski2023hyperparameter,mahajan2024empirical}. Here, we propose one additional solution by studying the covariate distribution balances, and we compare the empirical performance of those criteria. 
 
\subsection{Definitions of selection criteria}

\paragraph{Nuisance model performance.}
One straightforward approach is to use the performance of the supervised learning of the nuisance model as the selection criterion.  As first studied in \citet{athey2016recursive}, the generalization performance by performing a train-validation-test split is of vital importance. Here, we apply cross-validation (CV) since the sample size in our real application is relatively small. In our setting, the nuisance model is the propensity score model which learns the binary treatment assignment, and we choose several binary prediction metrics to quantify the performance of the propensity score model---$\ell_2$ norm of the prediction error ($\ell_2$ err.), Cross Entropy error (CE err.), and the area under the receiver operating characteristic curve (AUC). 

In our study, the AUC is implemented using \texttt{sklearn.metrics.auc} in Python, and $\ell_2$ err. as well as CE err. are defined as follows:

\begin{definition}[$\ell_2$ err.]
    Given binary samples $\{{a}_{i}, \ i = 1,\dots,n\}$ their predictions $\{\hat{a}_{i}, \ i = 1,\dots,n\}$, the $\ell_2$ norm of the prediction error is given by
    \[\ell_2 \ {\rm err.} \ = \frac{1}{n_\myt} \sum_{i=1}^n ({a}_{i} - \hat{a}_{i})^2.\]
\end{definition}

\begin{definition}[CE err.]
    Given binary samples $\{{a}_{i}, \ i = 1,\dots,n\}$ their predictions $\{\hat{a}_{i}, \ i = 1,\dots,n\}$, the Cross Entropy error is given by
    \[{\rm CE \ err.} \ = \frac{1}{n_\myt} \sum_{i=1}^n   {a}_i \log(\hat{a}_i) + (1 - {a}_i) \log(1 - \hat{a}_i) .\]
\end{definition}

\paragraph{Covariate balance.}
As the fitted propensity scores are used to balance the covariate distribution, one popular method to assess the goodness-of-fit in causal inference is to directly examine the ``balances'' of the covariate distribution of the re-weighted samples. One commonly used measure of the covariate distribution ``balances'' is the standardized mean difference (SMD) \citep{zhang2019balance}, which is essentially Cohen's \texttt{d} and can measure the discrepancy between two probability distributions; Alternatively, one can use the distance between two probability distributions to measure the balances and here we choose Maximum Mean Discrepancy (MMD).

One commonly used measure of the covariate distribution ``balance'' is the standardized mean difference (SMD) \citep{zhang2019balance}, which is essentially the Cohen's \texttt{d}:

\begin{definition}[Cohen's \texttt{d}]
    Given two sets of samples $\cA = \{a_{i}, \ i = 1,\dots,m\}$ and $\cB = \{b_{j}, \ j = 1,\dots,n\}$, the Cohen's \texttt{d} is defined as follows: 
\begin{equation*}
    \texttt{d}_{\rm Cohen}\big(\cA,\cB\big) =\left(\bar{a}-\bar{b}\right) \Bigg/ {\sqrt{\frac{S_{\cA}^2+S_{\cB}^2}{2}}},
\end{equation*}
where \(\bar{a} = \sum_{i=1}^m a_i/m, \ \bar{b} = \sum_{i=1}^n b_j/n\), and the pooled sample variance can be calculated as:
\begin{equation*}
    \begin{split}
        &S_{\cA} = \frac{1}{m} \sum_{i=1}^m \left(a_i - \bar{a}\right)^2, \\
        &S_{\cB} = \frac{1}{n} \sum_{j=1}^n \left(b_j - \bar{b}\right)^2.
    \end{split}
\end{equation*}
\end{definition}

When Cohen's \texttt{d}'s absolute value is close to zero, the standardized means of two distributions are similar to each other, \ie, the covariates' distributions are balanced.
In our multivariate setting, we can simply use the average of the absolute Cohen's \texttt{d}'s as the selection criterion, \ie,
\[\text{SMD} = \frac{1}{d}\sum_{j=1}^{d} \Big|\texttt{d}_{\rm Cohen}\big(\{\boldsymbol{x}(j): \boldsymbol{x} \in \cD_{\boldsymbol{x},\rm trt}\}, \{\boldsymbol{x}(j): \boldsymbol{x} \in \cD_{\boldsymbol{x},\rm ctrl}\}\big)\Big|, \]
where $\boldsymbol{x}(j)$ is the $j$-th element of the vector $\boldsymbol{x} \in \RR^d$, and
\begin{equation*}
    \begin{split}
        &\cD_{\boldsymbol{x},\rm trt} = \left\{ \frac{\boldsymbol{x}_{i,\myt}}{g(\boldsymbol{x}_{i,\myt}^\T \hat \beta_\myt)}: {a}_{i,\myt} = 1\right\}, \\
        &\cD_{\boldsymbol{x},\rm ctrl} = \left\{ \frac{\boldsymbol{x}_{i,\myt}}{1-g(\boldsymbol{x}_{i,\myt}^\T \hat \beta_\myt)}: {a}_{i,\myt} = 0\right\}.
    \end{split}
\end{equation*}
The above metric tells us the average standardized mean difference after propensity score weighting, which we will refer to as the SMD in the following analysis.

As one can see, the goal is to quantitatively characterize how similar two empirical distributions are after propensity score weighting, and this is indeed extensively suited to two sample test problems. Here, we introduce one very popular non-parametric distance metric between distributions (\ie, two-sample test statistic), Maximum Mean Discrepancy \citep{gretton2012kernel}, as follows:

\begin{definition}[Maximum Mean Discrepancy]
    Given two sets of samples $\cA = \{\boldsymbol{a}_{i}, \ i = 1,\dots,m\}$ and $\cB = \{\boldsymbol{b}_{j}, \ j = 1,\dots,n\}$, an unbiased estimator of MMD can be obtained via U-statistics as follows: 
\begin{equation*}
\hat{\text{MMD}}^2(\mathcal{A}, \mathcal{B}) = \frac{1}{m(m-1)}\sum_{i=1}^{m} \sum_{j\neq i}^{m} k(\boldsymbol{a}_i, \boldsymbol{a}_j) - \frac{2}{mn}\sum_{i=1}^{m} \sum_{j=1}^{n} k(\boldsymbol{a}_i, \boldsymbol{b}_j) + \frac{1}{n(n-1)}\sum_{i=1}^{n} \sum_{j\neq i}^{n} k(\boldsymbol{b}_i, \boldsymbol{b}_j).
\end{equation*}
\end{definition}
Our MMD-based selection criterion (referred to as MMD for brevity) is then given as:
\[\text{MMD} = \hat{\text{MMD}}^2\left(\cD_{\boldsymbol{x},\rm trt},\cD_{\boldsymbol{x},\rm ctrl}\right).\]
In the above definition,
$k$ is the user-specified kernel function, \ie, \(k(\cdot,\cdot): \cX \times \cX \rightarrow \RR.\)
Commonly used kernel functions include 
Gaussian radial basis function $k(\boldsymbol{x}_1,\boldsymbol{x}_2) = \exp \{-{\norm{\boldsymbol{x}_1-\boldsymbol{x}_2}_2^2}/{r^2}\},$ where $\norm{\cdot}_2$ is the vector $\ell_2$ norm and ${{r}} > 0$ is the bandwidth parameter; The bandwidth is typically chosen using median heuristic.

\subsection{Numerical simulation results}

\paragraph{Configuration.}
As the counterfactuals are unobserved in practice, we conduct numerical simulation, in which we know the ground truth causal effect, to compare the effectiveness of the aforementioned selection criteria. We consider a $d=50, s=2$ example where we have $n_\myt = 100$ target domain samples and $n_\mys = 2000$ source domain samples.
The covariates are taken as the absolute values of normally distributed random numbers with mean $0$ and standard deviation $1$. The source domain propensity score (PS) model parameters are also absolute values of normally distributed random numbers with mean $0$ and standard deviation $0.1$. In the data generating process, the ``unknown'' true PS model is GLM with exponential link function whereas our specified model considers GLM with sigmoid link function, \ie, we consider model mismatch in our simulation. We take absolute values since the link function, \ie, the exponential function $g(x) = 1 - \exp\{-x\}$, is supported on $x \in (0,\infty)$. The $s$-sparse difference has magnitude $0.2$. The ground truth ACE is $\tau_\mys = 5$ in the source domain and $\tau_\myt = 3$ in the target domain.
Our specified model is GLM with sigmoid function and we fit vanilla logistic regression using source domain data for the rough estimation step. Afterwards, we perform gradient descent to learn the sparse difference with $\ell_1$ regularization; The total number of iterations is $20000$, the initial learning rate (lr) is $0.001$ and it decays by $1\%$ (\ie, lr $\leftarrow 0.99$ lr) every $1000$ iterations.

\begin{wrapfigure}{r}{0.5\textwidth}
\vspace{-0.2in}
\centerline{
\includegraphics[width = .4\textwidth]{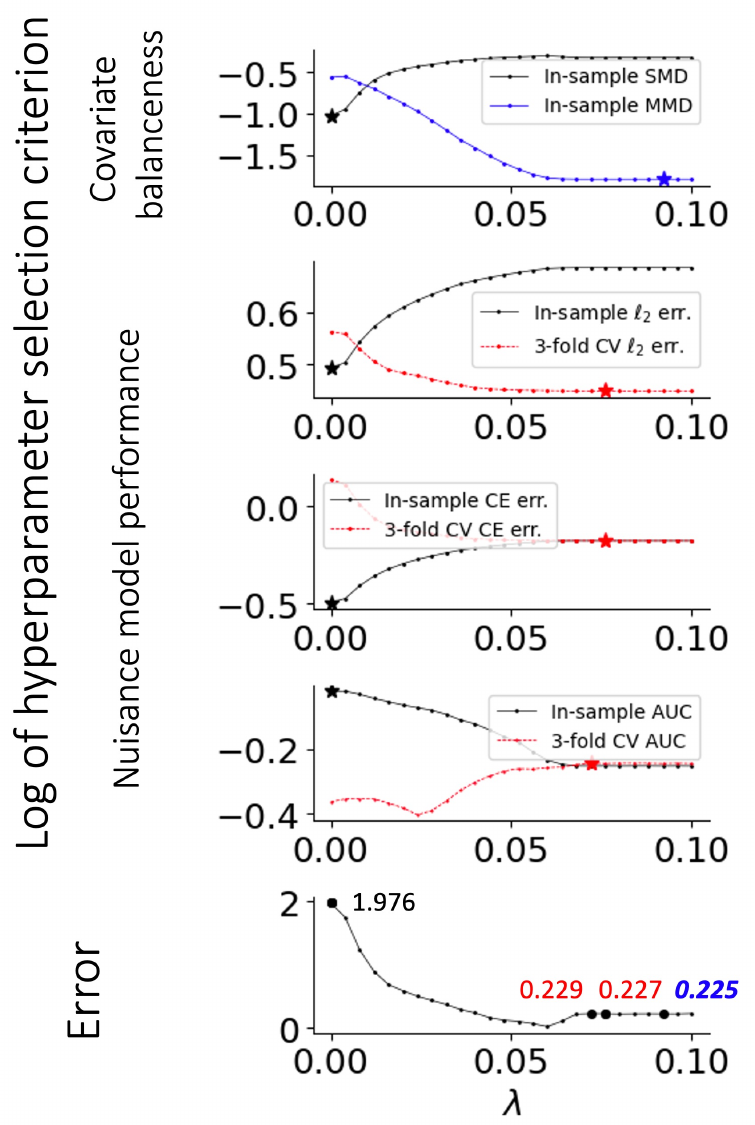} 
}
\vspace{-0.1in}
\caption{\small Comparison of different hyperparameter selection criteria: We use ``$\star$'' to denote the selected hyperparameter in the first four rows; The corresponding $\ell_1$-TCL estimation errors are highlighted with ``$\bullet$'' in the last row. We can observe that in-sample MMD and all cross-validation nuisance model performance metrics can select $\lambda$'s that output similar and accurate causal effect estimates.}
\label{fig:simu}
\vspace{-0.05in}
\end{wrapfigure}
\paragraph{Results.}
If we estimate the causal effect using the target domain data only, the IPW estimate will have a very large error (estimation error is $1.45$ whereas the ground truth causal effect is $\tau_\myt = 3$). 
We perform the aforementioned $\ell_1$-regularized transfer learning and plot how the aforementioned criteria, as well as the resulting $\ell_1$-TCL estimation errors, vary with different $\lambda$ in Figure~\ref{fig:simu}.

Surprisingly, the SMD considered in \citet{zhang2019balance} does not perform well in our set-up, and it may require more sophisticated aggregation (than the simple average) of the Cohen's \texttt{d}'s. The observation that in-sample nuisance model performances are rather poor agrees with existing literature \citep{athey2016recursive,curth2023search,machlanski2023hyperparameter}. Importantly, we find that the MMD criterion has a similar performance to the CV nuisance model performance criteria; Indeed, it performs the best in this specific example.

One drawback of all aforementioned criteria is that they do not involve outcome variables, and therefore, they cannot be used to test whether or not the heterogeneous causal effect assumption holds. As a result, performing subgroup analysis and transfer learning among subgroups might be the most robust approach that we can take. In our numerical example, IPW estimation on the entire dataset (\ie, both domains combined) yields a point estimate of $4.91$, which is very close to the source domain true causal effect $\tau_\mys = 5$. Additionally, we consider the inference using only the target domain dataset, and the IPW estimate is $4.45$, which is still less than satisfactory compared with the $\ell_1$-TCL estimate.

\section{Electricity Service Unfairness Assessment Using City-Level Outage Data}
In this section, after introducing ``what-is'' (\ie, our \cupo) and ``how-to'' (\ie, our $\ell_1$-TCL) parts of this work, we are ready to present the experimental results on unfairness assessment in electricity service systems. We compare our $\ell_1$-TCL approach with baseline methodologies such as naïve correlation studies and the vanilla IPW estimator. We provide point estimates, integrating hyperparameter selection, along with Bootstrap uncertainty quantification findings. Further details, such as dataset descriptions and additional results on hyperparameter selection, are deferred to Appendix~\ref{appendix:real_exp} due to space limitations.

\subsection{Dataset}\label{sec:data}
We collect data from multiple sources, including city-level outage data that integrates information from all service territories \citep{Bryan2012,Campbell2012}, socio-demographic factors, and weather information: The outage data was collected for 351 cities in Massachusetts from the local government, spanning 2018/03/01 to 2018/03/31. The weather data was derived from the High-Resolution Rapid Refresh (HRRR) model, while demographic details were sourced from the American Community Survey (ACS) 5-Year survey conducted by the US Census Bureau. Additionally, land cover and utilization data were procured from the Massachusetts government's MassGIS Data: 2016 Land Cover/Land Use (LCLU) dataset.

Our real-data experiment examines the counterfactual influences of certain protected attributes (\ie, the ``treatment'' variable), including median income, elderly percentage, and total customer number, on the duration of power outages (\ie, the outcome variable), taking into account additional attributes, such as demographics, land cover, land use, and weather conditions. The outage duration is measured using the SAIDI---System Average Interruption Duration Index---which is a common measure of the average outage duration experienced by each customer over a specific timeframe (in this case, quantified as outage minutes per month); See its formal definition in \eqref{eq:saidi} in the Appendix. The weather factors include wind speed ($m/s$) and precipitable water ($kg/m^2$).

To account for potential heterogeneity, outage records are divided into two subgroups: severe and normal weather conditions. An outage was classified under severe weather if either the peak wind speed exceeds $19 m/s$ or the maximum precipitable water surpasses $16 kg/m^2$. SAIDI values are computed separately for each weather category, facilitating a dual-dataset approach for the transfer counterfactual learning. For a comprehensive understanding of SAIDI, demographic aspects, as well as land cover and usage factors employed in this analysis, refer to Appendix~\ref{appendix:realdata}.

\subsection{Point estimate}

\subsubsection{Unfairness assessment w.r.t. wealthiness}
We first designate median income as the protected attribute. This attribute is dichotomized based on whether it surpasses the $80\%$ percentile threshold across the entire population. This process yields a ``treatment indicator'', categorizing a city as economically disadvantaged relative to the $80\%$ percentile benchmark of median income across all cities.

\paragraph{Naïve approaches.} 
Our initial analysis uses naïve correlation analysis, revealing a correlation of \(-0.06\) between ``wealthiness indicator'' and power outage duration under severe weather conditions, and \(0.43\) under normal conditions; These results are slightly different than Figure~\ref{fig:counterfactual_illus}(a) as we binarize the median income here. When incorporating other attributes into a regression analysis, the ordinary least square (OLS) coefficient (coef.) for the protected attribute is $-92.15$ ($p$-val = 0.65) under severe weather conditions and $62.87$ ($p$-val = 0.61) under normal conditions, indicating a lack of significant evidence to reject the null hypothesis of zero coefficient in both scenarios. These two methods are referred to as the \emph{native approaches} in the following.

\paragraph{Counterfactual approach: vanilla IPW.} 
In our third baseline approach, we apply the \emph{(vanilla) IPW estimator without (w/o) TL} to adjust for the potential confounders in other attributes. Our results show similar patterns, with estimated ACEs of \(-460.79\) for the severe weather subgroup and \(68.03\) for the normal weather subgroup. These findings suggest that wealthier areas experienced longer power outages under normal weather conditions, a result that disagrees with existing literature and appears counterintuitive.

\begin{figure}[!htp]
\centerline{
\includegraphics[width = .9\textwidth]{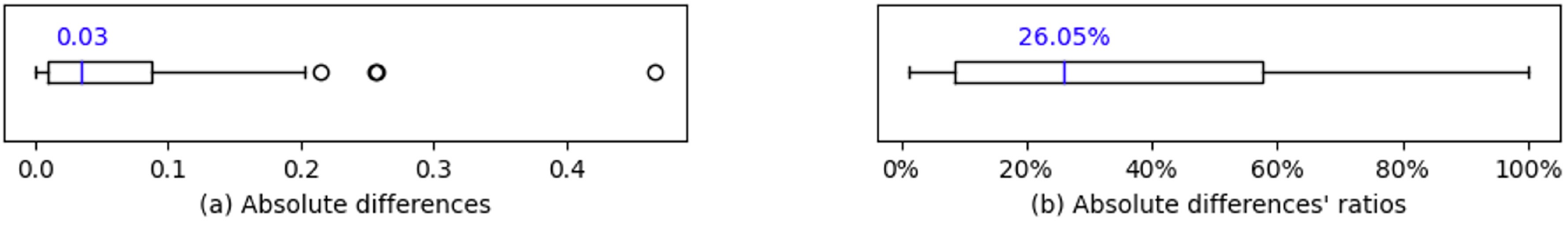} 
}
\caption{\small Box plots of the absolute values 
 in the difference of the estimated nuisance parameters, (\ie, \(\hat \beta_\myt - \hat \beta_\mys\)), and their ratios (\ie, the difference divided by \((\hat \beta_\myt + \hat \beta_\mys)/2\)), where \(\hat \beta_\myt\) and \(\hat \beta_\mys\) are estimated using $\ell_1$ regularized logistic regression (where wealthiness indicator is the response variable and other attributes are the predictors). Although most of the raw estimated differences are close to zero (as shown in sub-figure (a)), the estimated nuisance parameters themselves could also be close to zero. Therefore, we also report the ratios in sub-figure (b), which shows that the estimated differences are mostly close to zero relative to their raw estimates, which supports Assumption~\eqref{assumption}.}
\label{fig:realExpMADelta}
\end{figure}

\paragraph{Counterfactual approach: proposed $\ell_1$-TCL.} 
To improve the estimation accuracy, we employ the $\ell_1$-TCL method we developed. Initially, we study the validity of the ``sparse nuisance parameters' difference'' assumption, \ie, Assumption~\eqref{assumption}. For this, we utilize a straightforward $\ell_1$ regularized logistic regression to roughly estimate the nuisance parameters for both domains and deduce their difference as the estimated difference. The magnitudes of the absolute differences, along with their proportions relative to the raw nuisance parameter estimates, are depicted in the box plots in Figure~\ref{fig:realExpMADelta}.

Supported by the evidence in Figure~\ref{fig:realExpMADelta}, we proceed with our $\ell_1$-TCL coupled with various criteria. The estimated ACEs using $\ell_1$-TCL are documented in Table~\ref{table:MA}; Here, the symbol ``$\boldsymbol{\uparrow}$'' denotes our binarization criterion, specifically, surpassing the $80\%$ percentile threshold. Notably, our $\ell_1$-TCL approach in Table~\ref{table:MA} utilizes in-sample MMD as the criterion for hyperparameter selection; Additional results with other criteria are deferred to Figure~\ref{fig:realExpMAHypSelection} in Appendix~\ref{sec:hyp_selection}.

The results showcased in Table~\ref{table:MA} reveal that our $\ell_1$-TCL method deduces causal effect estimates of $-984.84$ under severe weather conditions and $-881.88$ for normal weather scenarios. This consistently suggests significant discrimination, marked by extended power outages, particularly in economically disadvantaged areas---a situation that further deteriorates under severe weather conditions. This finding is further corroborated by the uncertainty quantification outcomes, which will be discussed subsequently.

\begin{table}[!htp]
    \centering
    \caption{\small Point estimates of the ACE. We quantize the estimated ACE \(\hat \tau\) as follows: ``\newmm'' if $\hat \tau \leq -500$, ``\newnewm'' if $-500 < \hat \tau < -100$, ``\newnewn'' if $-100 \leq \hat \tau \leq 100$, ``\newnewp'' if $\hat \tau > 100$. The results highlight the impact of TL in producing outcomes that more closely align with both established literature and intuitive understanding, in contrast to estimates derived without leveraging knowledge from the source domain (\ie, w/o TL).}
    \label{table:MA}
    \vspace{.1in}
    \resizebox{.95\textwidth}{!}{
        \begin{tabular}{lcccccccc}
         \toprule
         & \multicolumn{2}{c}{{Median income}$_{(\uparrow)}$} & \multicolumn{2}{c}{{Elderly percentage}$_{(\uparrow)}$}& \multicolumn{2}{c}{{Customer number}$_{(\uparrow)}$}\\
        Weather  & Severe & Normal  & Severe & Normal & Severe & Normal \\
\cmidrule(l){2-3} \cmidrule(l){4-5} \cmidrule(l){6-7} 
$\ell_1$-TCL & \newmm \lowernum{-984.84} & \newmm \lowernum{-881.88} & \newp \lowernum{420.74} & \newp \lowernum{255.76} & \newmm \lowernum{-616.26} & \newn \lowernum{34.55}\\
W/o TL & \newm \lowernum{-460.79}  & \newn \lowernum{68.03} & \newm \lowernum{-253.89} & \newm \lowernum{-146.62} & \newmm \lowernum{-557.00} & \newm \lowernum{-141.18}\\
         \bottomrule
        \end{tabular}
        }
\end{table}

\subsubsection{Unfairness assessment for additional protected attributes}
Our study broadens to encompass other protected attributes---the elderly percentage and the customer number. We adopt a similar approach for categorizing these protected attributes into treatment indicators, setting the binarization threshold at the \(80\%\) percentile of the entire population. The point estimates of ACE using $\ell_1$-TCL, coupled with the MMD criterion, are presented in Table~\ref{table:MA}.

Observations from Table~\ref{table:MA} consistently indicate that the vanilla counterfactual approach (\ie, IPW w/o TL) tends to yield incongruent results, counterintuitively suggesting that regions with a larger elderly populace face fewer power outages, a pattern that persists across various weather conditions. 
In contrast, results from our $\ell_1$-TCL portray a more realistic scenario, showing extended power restoration times in areas with a higher percentage of elderly residents. Moreover, this disparity is notably more accentuated under severe weather conditions. These findings hint at a possible indirect causal relationship between the elderly population and power outage durations, potentially mediated by some unobserved factors. 
Interestingly, $\ell_1$-TCL also reveals that extended power outages in less densely populated areas only exist under severe weather conditions, which is an explainable discrimination---This could be interpreted as a strategic prioritization by power companies to restore service to the majority of their customers swiftly under extreme conditions, representing a form of rational (or explainable) discrimination.

\begin{figure}
\centering
\vspace{-0.1in}
      \includegraphics[width = .94\textwidth]{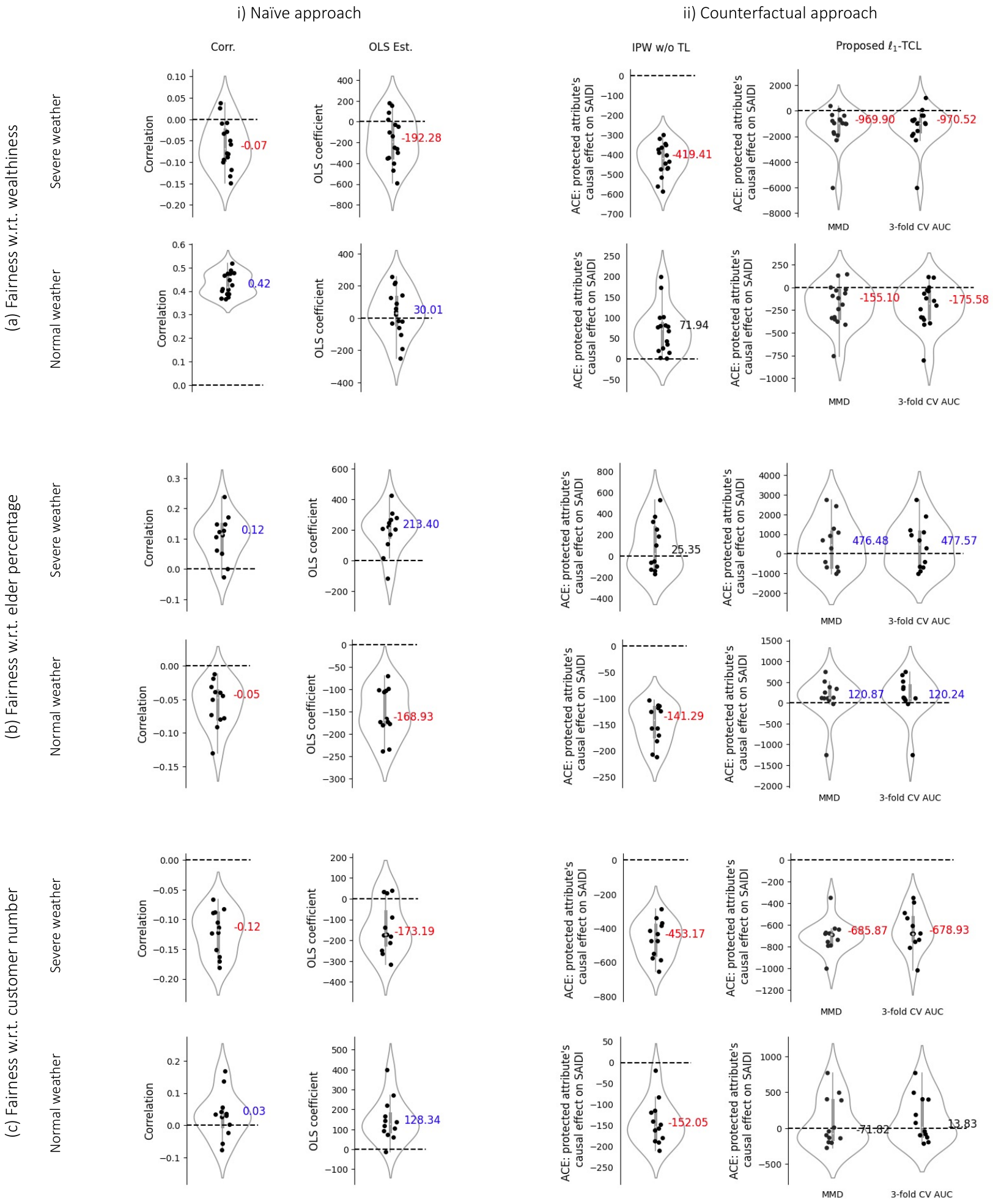}
  \vspace{-0.1in}
  \caption{\small Bootstrap UQ visualization for unfairness assessment w.r.t. protected attributes (as detailed in the sub-captions on the left). In each violin plot, the median is highlighted: For the naïve approach, {\color{blue} blue} signifies a positive effect, while {\color{red} red} denotes a negative effect; For counterfactual approaches, {\color{blue} blue} (median ACE $>-100$) represents a positive causal effect, {\color{red} red} (median ACE $<-100$) indicates a negative effect, and black ($-100 \leq$ median ACE $\leq 100$) suggests a neutral effect. Notably, naïve methods oftentimes output results lacking statistical significance; Both naïve methods and vanilla IPW (\ie, w/o TL) are prone to produce outcomes that contradict both established literature and intuitive reasoning; In contrast, our proposed $\ell_1$-TCL constantly (irrespective of the protected attribute or hyperparameter selection criterion) delivers findings that align with prevailing research.}\label{fig:realExpMABP}
\end{figure}

\subsection{Uncertainty quantification}\label{sec:uq}
Beyond the point estimates previously discussed, we also perform Bootstrap uncertainty quantification (UQ) for the results. The violin plots depicting these UQ results are illustrated in Figure~\ref{fig:realExpMABP}. Our $\ell_1$-TCL incorporates both in-sample MMD criterion and cross-validation (CV) nuisance model performance (in this case, the area under ROC curve (AUC)) for hyperparameter selection. Further details, including Bootstrap configurations and $\ell_1$-TCL results with additional criteria, can be found in Appendix~\ref{appendix:BPmore}. 

The UQ results presented in Figure~\ref{fig:realExpMABP} align with our earlier point estimates, reinforcing the reliability of our findings. Notably, our analysis suggests that normal weather conditions present a more challenging scenario for baseline approaches, such as correlation studies and vanilla IPW, to accurately capture the relationship between socio-demographic attributes and power outage duration. This challenge may stem from the reduced frequency of power outages under normal conditions, potentially obscuring the differences in outage durations across various socio-demographic groups.
Moreover, Figure~\ref{fig:realExpMABP} reveals that baseline methodologies, such as correlation studies and the vanilla IPW estimator, occasionally produce counterintuitive findings that disagree with existing literature. These methods have indicated, for instance, an extended power outage duration in wealthy neighborhoods under normal weather conditions. This is inconsistent with reasonable expectations of discrimination. Despite setting a threshold where ACE with an absolute value less than $100$ is considered to have ``no effect,'' the vanilla IPW estimator still shows positive effects across all Bootstrap trials under normal weather conditions, as shown in Figure~\ref{fig:realExpMABP}(a). 
Furthermore, some baseline approaches yield statistically insignificant results, limiting the conclusions that can be drawn. Examples of this include the correlation study under normal weather conditions (Figure~\ref{fig:realExpMABP}(c)), the OLS estimator (Figure~\ref{fig:realExpMABP}(a)), and the IPW estimator under severe weather conditions (Figure~\ref{fig:realExpMABP}(b)). In contrast, our $\ell_1$-TCL method consistently produces more plausible results, aligning with both our point estimate findings and existing literature:
Similar incongruous findings regarding discrimination against elderly populations and total customer number under normal weather conditions are observed in the UQ results.

\section{Conclusion and Discussion} 

This work shows that counterfactual unfairness can be rigorously assessed using average causal effect under Rubin's Potential Outcome framework. The introduction of $\ell_1$-TCL marks an advancement in addressing the challenge of data scarcity in causal inference. The adoption of $\ell_1$-TCL to assess the energy injustice is more than a mere methodological selection---It is a paradigm shift in the analytical lens through which the complex relationship between socio-demographic variables and power outage durations is understood. At the same time, our $\ell_1$-TCL is not limited to energy justice---It can be generalized to many applications facing the confounding bias, model heterogeneity, and data scarcity issues, which are commonly seen in practice.

Our transfer counterfactual analysis shows that our $\ell_1$-TCL consistently provides more reasonable results compared to the various baseline approaches. Importantly, it highlights prolonged power outages in areas characterized by both low-income and a higher percentage of elderly residents. This phenomenon occurs under both severe and normal weather conditions, indicating that the economically disadvantaged and elderly-populated areas may face persistent challenges in maintaining power supply reliability. To make it worse, these areas may have increased vulnerability (\ie, longer outage periods) during severe weather events. 
This suggests a need for targeted infrastructure improvements and support in economically disadvantaged regions to mitigate unfairness observed in the existing power service systems.

While this work offers plausible insights into the unfairness in the power service systems, we acknowledge the limitations inherent in the Ignorability Assumption \eqref{assumption:ignorability}, a cornerstone yet untestable assumption in causal inference literature. For instance, in our motivating example, critical factors such as infrastructure quality and proximity to major electricity service sites could influence the assessment results. Although such data is not available in our current study, and we posit that the relatively small area of Massachusetts might mitigate the impact of infrastructure availability, this still represents a notable gap. We attempt to mitigate this by incorporating as much available information as possible to account for potential confounders comprehensively. Furthermore, we recognize the absence of ground truth for external validation in our real example, and therefore, we will release our implementation post-review to facilitate robust external validation, ideally on benchmark datasets with established causal relationships.
\section*{Acknowledgment}

The work of Song Wei and Yao Xie is partially supported by an NSF CAREER CCF-1650913, and NSF DMS-2134037, CMMI-2015787, CMMI-2112533, DMS-1938106, DMS-1830210.

\bibliographystyle{plainnat}
\bibliography{ref}

\newpage

\appendices

\addcontentsline{toc}{section}{Appendix} 
\part{\centering \LARGE Appendix} 

\topskip0pt

\parttoc 


\section{Theoretical Analysis on Estimation Error Bounds}\label{appendix:TLIPW}

We first establish a non-asymptotic upper bound on the error for the $\ell_1$ regularized TL estimator applied to propensity score (PS) models. After this, we will incorporate it into the error bound for the unbiased IPW estimator, as delineated in eq.~\eqref{eq:ipw}. This will lead us to the final performance guarantee for the $\ell_1$-TCL estimator. Our method of proof aligns closely with the approach taken by \citet{bastani2021predicting,wei2023transfer}. We provide the proof here mainly for the sake of completeness.

\subsection{Preliminaries} 

The notations used here follow standard conventions. $\norm{\cdot}_p$ denotes the vector $\ell_p$ norm. 
For asymptotic notations: 
$f(n) = o(g(n))$ or $g(n) \gg f(n)$ means for all $c > 0$ there exists $k > 0$ such that $0 \leq f(n) < cg(n)$ for all $n \geq k$;
$f(n) =  \cO(g(n))$ means there exist positive constants $c$ and $k$, such that $0 \leq f(n) \leq cg(n)$ for all $n \geq k$.

\subsection{Guarantee for PS model estimation with knowledge transfer}
We will start by setting the foundational assumptions:

\begin{assumption}\label{A0}
    The covariates in both the target and source domains are uniformly bounded, meaning there exists a constant $M_X > 0$ such that for all covariates $\boldsymbol{x}_{i, \myt}$ in the target domain and $\boldsymbol x_{i,\mys}$ in the source domain, their infinity norms are bounded by $M_X$, i.e., $\norm{\boldsymbol{x}_{i, \myt}}_{\infty} \leq M_X$ for $i = 1,\dots,n_{\myt},$ and similarly, $\norm{\boldsymbol x_{i,\mys}}_{\infty} \leq M_X$ for $i = 1,\dots,n_\mys$. 
\end{assumption}

This assumption slightly diverges from the ``standardized design matrix'' assumption in \citet{bastani2021predicting}, which necessitates the squared $F$-norms of design matrices $(\boldsymbol{x}_{1, \myt},\dots,\boldsymbol{x}_{n_{\myt}, \myt})^\T$ and $(\boldsymbol x_{1, \mys},\dots,\boldsymbol x_{n_{\mys}, \mys})^\T$ to be $n_{\myt}$ and $n_{\mys}$, respectively. Nonetheless, we will later demonstrate that they fulfill a similar role when we prove Lemma~\ref{lma:TL-beta}.

We define the sample covariance matrices as:
\begin{equation}\label{eq:cov}
    \Sigma_\myt = \frac{1}{n_{\myt}} \sum_{i = 1}^{n_{\myt}} \boldsymbol{x}_{i, \myt} \boldsymbol{x}_{i, \myt}^\T \in \RR^{n_{\myt} \times n_{\myt}}, \quad \Sigma_\mys = \frac{1}{n_\mys} \sum_{i = 1}^{n_\mys} \boldsymbol x_{i,\mys} \boldsymbol x_{i,\mys}^\T \in \RR^{n_\mys \times n_\mys}.
\end{equation}

\begin{assumption}\label{A1}
We posit that the source domain sample covariance matrix $\Sigma_\mys$ possesses a non-negative minimum eigenvalue denoted by $\psi > 0$, ensuring $\Sigma_\mys$ is positive-definite.
\end{assumption}

Assumption~\ref{A1} is critical as it guarantees the feasibility of accurately recovering $\beta_\mys$ through Maximum Likelihood Estimation (MLE) from the source domain data. This requirement is typically mild, particularly when the condition $n_\mys > d$ is met, a scenario that aligns with the regime we are considering in \eqref{eq:regime}.

\begin{definition}[Compatibility Condition \citep{bastani2021predicting}]
The compatibility condition with a constant $\phi > 0$ is satisfied for a subset of indices $\cI \subset \{1,\dots,d\}$ and a matrix $\Sigma \in \RR^{d \times d}$ if the following inequality holds for all vectors $u \in \RR^d$ that meet the criterion $\norm{u_{\cI^{\rm c}}}_1 \leq 3 \norm{u_{\cI}}_1$:
\[ u^\T \Sigma u \geq  \frac{\phi^2}{\# \cI} \norm{u_{\cI}}_1^2,\]
where $\#$ denotes the number of elements in a set, superscript $^{\rm c}$ represents the set complement, and $u_{\cI}$ refers to a vector whose $j$-th elements are $u_j$, that is, the $j$-th element of vector $u$, if $j$ is in the index set $\cI$, and zero otherwise.
\end{definition}

A standard assumption in high-dimensional Lasso literature is:

\begin{assumption}\label{A2}
The index set $\cI = \operatorname{supp}(\Delta_\beta)$ \eqref{eq:beta_diff} and target domain sample covariance matrix $\Sigma_\myt$ \eqref{eq:cov} meet the above compatibility condition with constant $\phi > 0$.
\end{assumption}

This assumption is pivotal for ensuring the identifiability of $\Delta_\beta$. While it is inherently met when the target domain sample covariance is positive-definite, in scenarios where $n < d$, the target domain sample covariance becomes rank-deficient. In such instances, Assumption~\ref{A2} becomes indispensable for affirming the identifiability of $\Delta_\beta$.

The last assumption before we present the first technical result is standard in GLM literature, and it is automatically satisfied when the link function $G'(\cdot) = g(\cdot)$ is linear, \ie, $g(x) = x$ with domain $x \in [0,1]$. 

\begin{assumption}\label{A3}
    The function $G(\cdot)$ is strongly convex with $\gamma > 0$; Specifically, for any $w_1,w_2$ within its domain, the following inequality holds:
    \[G(w_1) - G(w_2) \geq G'(w_2) (w_1 - w_2) + \gamma \frac{(w_1 - w_2)^2}{2}.\]
\end{assumption}

Now, we are ready to present the recovery guarantee for the $\ell_1$ regularized TL for the PS model nuisance parameters:

\begin{lemma}[Transferable Guarantee for PS Model]\label{lma:TL-beta}
Under Assumptions~\ref{A0}, \ref{A1}, \ref{A2}, and \ref{A3}, and presuming the propensity score model is accurately specified with the difference $\Delta_\beta$ \eqref{eq:beta_diff} being $s$-sparse, the estimator $\hat \beta_\myt^{\ \ell_1\rm{TCL}}$, equipped with a regularization strength hyperparameter $\lambda > 0$, satisfies the following probabilistic upper bound:
\begin{equation}\label{eq:upper_hatbeta}
\begin{split}
    \PP\left(\left\|\hat \beta_\myt^{\ \ell_1\rm{TCL}} -\beta_\myt \right\|_1 \geq \frac{5 \lambda}{\gamma}\left(\frac{1}{8 \psi^2}+\frac{1}{\psi}+\frac{s}{\phi^2}\right)\right) \leq  2 d \exp \left(-\frac{2\lambda^2 n_\myt}{125 M_X^2}\right)+2 d \exp \left(-\frac{2\lambda^2 n_\mys}{ 5 d^2 M_X^2}\right).
\end{split}
\end{equation}
\end{lemma}

This lemma provides a probabilistic guarantee on the performance of the $\ell_1$ regularized TL estimator for the PS model, ensuring its reliability under the specified conditions. In the forthcoming subsection, we utilize the derived error bound to quantify the deviation in the estimated propensity scores, specifically, the magnitude of $|g(\boldsymbol{x}_{i, \myt}^\T \beta_\myt) - g(\boldsymbol{x}_{i, \myt}^\T \hat \beta_\myt^{\ \ell_1\rm{TCL}})|$. This involves the application of Hölder's inequality, yielding:
\[\boldsymbol{x}_{i, \myt}^\T (\beta_\myt - \hat \beta_\myt^{\ \ell_1\rm{TCL}}) \leq |\boldsymbol{x}_{i, \myt}^\T (\beta_\myt - \hat \beta_\myt^{\ \ell_1\rm{TCL}})| \leq \|\boldsymbol{x}_{i, \myt}\|_{p_1} \|\beta_\myt - \hat \beta_\myt^{\ \ell_1\rm{TCL}}\|_{p_2},\]
where $1/p_1+1/p_2 = 1, \ p_1, p_2 \geq 1$. Common pairings for $(p_1,p_2)$ include $(2,2)$ and $(\infty,1)$. Despite the potential application of the Restricted Eigenvalue Condition \citep{candes2007dantzig,bickel2009simultaneous,meinshausen2009lasso,van2009conditions} to bound $\|\beta_\myt - \hat \beta_\myt^{\ \ell_1\rm{TCL}}\|_{2}$, which scales with $\sqrt{s}$ instead of $s$, under Assumption~\ref{A0}, $\|\boldsymbol{x}_{i, \myt}\|_{2}$ scales with $\sqrt{n_\myt}$, typically overshadowing the sparsity term in our considered regime \eqref{eq:regime}. Consequently, the overall upper bound on the ACE estimate degrades to $\cO(\sqrt{sn_\myt \log d/n_\myt})$, a deterioration from $\cO(s \sqrt{\log d/n_\myt})$, as will be elaborated below. This delineates our rationale for employing the Compatibility Condition to obtain the $\ell_1$ error bound for the estimated nuisance parameters, instead of resorting to the Restricted Eigenvalue Condition for the $\ell_2$ error bound.


\subsection{Guarantee for the plug-in $\ell_1$-TCL estimator}
To bound the absolute estimation error, \ie, $|\hat \tau_\myt^{\ \ell_1\rm{TCL}} - \tau|$, where we abuse the notation \(\tau\) to denote the ground truth ACE in the target domain, we additionally impose the following mild technical assumptions:

\begin{assumption}\label{A4}
    The target domain outcomes are uniformly bounded, \ie, there exists $M_Y > 0$ such that  $|y_{i, \myt}| \leq M_Y,  i = 1,\dots,n_\myt$. 
\end{assumption}


\begin{assumption}\label{A5}
    The propensity scores evaluated on the target domain data are bounded away from zero and one. Specifically, we assume there exists $0 < m_g < 1/2$ such that \[m_g \leq g(\boldsymbol{x}_{i, \myt}^\T \beta_\myt) \leq 1 - m_g, \quad i = 1,\dots,n_\myt.\]
\end{assumption}

Assumption~\ref{A5} is a conventional prerequisite in the proof of the theoretical guarantee of the IPW estimator. This assumption underpins classic asymptotic analysis concerning the consistency and asymptotic normality of the IPW estimator, as elaborated in seminal works by \citet{wooldridge2002inverse,wooldridge2007inverse} (specifically, refer to Theorems 3.1 and 4.1 in \citep{wooldridge2002inverse}). Utilizing Hoeffding's inequality, we can establish the following concentration result:

\begin{lemma}\label{lma:IPW}
Under Assumptions~\ref{A4} and \ref{A5}, for any $t > 0$, the following holds:
\begin{equation}\label{eq:upper_tau_1}
\begin{split}
    &\PP\left(\left|\frac{1}{n_\myt} \sum_{i = 1}^{n_\myt} \frac{{a}_{i} y_{i, \myt}}{g(\boldsymbol{x}_{i, \myt}^{\T} \beta_\myt )} - \frac{(1-{a}_{i}) y_{i, \myt}}{1-g(\boldsymbol{x}_{i, \myt}^{\T} \beta_\myt )} - \tau \right| \geq t\right) \leq 4  \exp \left(-\frac{m_g^2 t^2 n_\myt}{8 M_Y^2}\right).
\end{split}
\end{equation}
\end{lemma}

Before presenting the non-asymptotic guarantee for the $\ell_1$-TCL estimator, we impose the last technical assumption, which simplifies our derivation,
\begin{assumption}\label{Alip}
    The link function $g(\cdot)$ is $L$-Lipschitz with constant $L > 0$, \ie, for $x_1, x_2$ in its domain the following inequality holds: $|g(x_1) - g(x_2)| \leq L|x_1 - x_2|$.
\end{assumption}

Finally, with the help of the above lemmas, we can establish the non-asymptotic upper bound on the absolute estimation error of $\hat \tau_\myt^{\ \ell_1\rm{TCL}}$ as follows:
\begin{theorem}[Non-asymptotic recovery guarantee for $\hat \tau_\myt^{\ \ell_1\rm{TCL}}$ \eqref{eq:TLIPW}]\label{thm}
Under Assumptions~\ref{A0}, \ref{A1}, \ref{A2}, \ref{A3}, \ref{A4}, \ref{A5} and \ref{Alip}, for any constant $\delta > 0$, if the propensity score model is correctly specified and the difference $\Delta_\beta$ \eqref{eq:beta_diff} is $s$-sparse, as $n_\myt, n_\mys \rightarrow \infty$, suppose \eqref{eq:regime} holds, \ie,
\begin{equation*}
    s \sqrt{\frac{\log d}{n_\myt}}= o(1), \quad d \sqrt{\frac{n_\myt}{n_\mys}} = \cO(1),
\end{equation*}
we take $\ell_1$ regularization strength parameter to be
\begin{equation}\label{eq:lambda_1}
    \lambda = \sqrt{\frac{5 M_X^2 \log(6n_\myt d)}{2n_\myt}\max\left\{25, \frac{n_\myt d^2}{n_\mys}\right\}},
\end{equation}
and we will have
\begin{equation}\label{eq:main}
    \begin{split}
        \PP\Bigg(\left| \hat \tau_\myt^{\ \ell_1\rm{TCL}} - \tau \right| \leq (1+\delta) \left( C_1 s \sqrt{\frac{\log n_\myt + \log d}{n_\myt}\max\left\{1, \frac{n_\myt d^2}{25 n_\mys}\right\}} + \frac{2 M_Y}{m_g} \sqrt{\frac{\log n_\myt}{n_\myt}}\right)&\Bigg)  \\  \geq 1 - &\frac{1}{n_\myt},
    \end{split}
\end{equation}
where constant $C_1 = C_1(M_X,M_Y,\psi,\phi;\gamma,m_g,L)$ is defined as:
\begin{equation*}
    C_1 = \frac{100\sqrt{5}M_X^2 M_Y L}{\sqrt{2}m_g^2 \gamma}\left(\frac{1}{8\psi^2} + \frac{1}{\psi} + \frac{1}{\phi^2}\right).
\end{equation*}
\end{theorem}

\subsection{Proofs}\label{appendix:proof_TLIPW}

\begin{proof}[Proof outline of Lemma~\ref{lma:TL-beta}]
This proof mostly follows the proof of Theorem 6 in \citet{bastani2021predicting}. The differences in our setting come from: \textbf{(i)} The Bernoulli r.v.s are sub-Gaussian with variance bounded by $1/4$, which implies
\[\EE[{A} - g(\boldsymbol{X}_{\myt}^\T \beta_\myt)] = 0, \quad \Var({A} - g(\boldsymbol{X}_{\myt}^\T \beta_\myt)) \leq 1/4 + 1 = 5/4.\]
We need to substitute the variance terms with this upper bound (\ie, $5/4$). 

\vspace{0.03in}

\noindent
\textbf{(ii)} By Assumption~\ref{A0}, we have
\[\sum_{i=1}^{n_\myt} (\boldsymbol{x}_{i, \myt})_j^2 \leq n_\myt M_X^2,\]
where $(\boldsymbol{x}_{i, \myt})_j$ denotes the $j$-th element in the vector $\boldsymbol{x}_{i, \myt}$.
This implies that $\sum_{i=1}^{n_\myt} ({a}_i - g(\boldsymbol{x}_{i, \myt}^\T \beta_\myt))(\boldsymbol{x}_{i, \myt})_j$ is ($\sqrt{5n_\myt}M_X/2$)-sub-Gaussian (cf. Lemma 16 in \citet{bastani2021predicting}).
Notice that this is different from ``$\sum_{i=1}^{n_\myt} (\boldsymbol{x}_{i, \myt})_j^2 = n_\myt$'' due to the ``normalized feature assumption'' in the proof of Lemma 4 \citet{bastani2021predicting}. Therefore, in addition to substituting the variance terms as mentioned in (i), we need to include the additional $M_X$ term due to different model assumptions. 
Lastly, we perform the same modification to Lemma 5 and its proof in \citet{bastani2021predicting}, and these lead to \eqref{eq:upper_hatbeta}. For complete details of the proof, we refer readers to Appendix C in \citet{bastani2021predicting}.
\end{proof}

\begin{proof}[Proof of Lemma~\ref{lma:IPW}]
For the correctly specified propensity score model, the IPW estimator is unbiased, as shown in eq.~\eqref{eq:IPW_unbiased}.
Notice that Assumptions~\ref{A4} and \ref{A5} ensures 
\[\left |\frac{{a}_{i} y_{i, \myt}}{g(\boldsymbol{x}_{i, \myt}^{\T} \beta_\myt )}\right| \leq \frac{M_Y}{m_g}.\]
By Hoeffding's inequality, we have
\begin{equation*}
\begin{split}
    &\PP\left(\left|\frac{1}{n_\myt} \sum_{i = 1}^{n_\myt} \frac{{a}_{i} y_{i, \myt}}{g(\boldsymbol{x}_{i, \myt}^{\T} \beta_\myt )}  - \EE[Y_{1}] \right| \geq t\right) \leq 2  \exp \left(-\frac{m_g^2 t^2 n_\myt}{2 M_Y^2}\right).
\end{split}
\end{equation*}
Here, we slightly abuse the notation and use \(Y_{1}\) to denote the potential outcome in the treatment group in the target domain.
Similarly, \(Y_{1}\) refers to the potential outcome in the control group in the target domain, and we have $
\EE\left[\frac{(1-{A}) Y}{1-e(\boldsymbol{X})}\right]=E[Y_0]$, which implies
\begin{equation*}
\begin{split}
    &\PP\left(\left|\frac{1}{n_\myt} \sum_{i = 1}^{n_\myt} \frac{(1-{a}_{i}) y_{i, \myt}}{1 - g(\boldsymbol{x}_{i, \myt}^{\T} \beta_\myt )}  - \EE[Y_0] \right| \geq t\right) \leq 2  \exp \left(-\frac{m_g^2 t^2 n_\myt}{2 M_Y^2}\right).
\end{split}
\end{equation*}

Recall that $\tau = \EE[Y_1] - \EE[Y_0]$, we have 
\begin{align*}
    &\PP\left(\left|\frac{1}{n_\myt} \sum_{i = 1}^{n_\myt} \frac{{a}_{i} y_{i, \myt}}{g(\boldsymbol{x}_{i, \myt}^{\T} \beta_\myt )} - \frac{(1-{a}_{i}) y_{i, \myt}}{1-g(\boldsymbol{x}_{i, \myt}^{\T} \beta_\myt )} - \tau \right| \geq t\right) \\
    \leq &\PP\left(\left|\frac{1}{n_\myt} \sum_{i = 1}^{n_\myt} \frac{{a}_{i} y_{i, \myt}}{g(\boldsymbol{x}_{i, \myt}^{\T} \beta_\myt )}  - \EE[Y_1] \right| \geq t/2\right) + \PP\left(\left|\frac{1}{n_\myt} \sum_{i = 1}^{n_\myt} \frac{(1-{a}_{i}) y_{i, \myt}}{1 - g(\boldsymbol{x}_{i, \myt}^{\T} \beta_\myt )}  - \EE[Y_0] \right| \geq t/2\right) 
 \\
    \leq & 4  \exp \left(-\frac{m_g^2 t^2 n_\myt}{8 M_Y^2}\right).
\end{align*}
We complete the proof.
\end{proof}
\begin{proof}[Proof of Theorem~\ref{thm}]
One one hand, plugging the regularization parameter choice \eqref{eq:lambda_1} into \eqref{eq:upper_hatbeta} yields:
\begin{equation}\label{eq:pf1}
    \PP\left(\left\|\hat \beta_\myt^{\ \ell_1\rm{TCL}} -\beta_\myt \right\|_1 \geq \frac{5 \lambda}{\gamma}\left(\frac{1}{8 \psi^2}+\frac{1}{\psi}+\frac{s}{\phi^2}\right)\right) \leq \frac{2}{3n_\myt}.
\end{equation}
On the other hand, by setting $t = \frac{2 M_Y}{m_g} \sqrt{\frac{\log(12n_\myt)}{n_\myt}}$ in eq.~\eqref{eq:upper_tau_1} we have
\begin{equation}\label{eq:pf2}
    \PP\left(\left|\frac{1}{n_\myt} \sum_{i = 1}^{n_\myt} \frac{{a}_{i} y_{i, \myt}}{g(\boldsymbol{x}_{i, \myt}^{\T} \beta_\myt )} - \frac{(1-{a}_{i}) y_{i, \myt}}{1-g(\boldsymbol{x}_{i, \myt}^{\T} \beta_\myt )} - \tau \right| \geq \frac{2 M_Y}{m_g} \sqrt{\frac{\log(12n_\myt)}{n_\myt}}\right) \leq \frac{1}{3n_\myt}.
\end{equation}

Due to Assumptions~\ref{A4} and \ref{A5}, we have
\begin{equation}
    \begin{split}\label{eq:pf3}
        \left|\frac{1}{n_\myt} \sum_{i = 1}^{n_\myt} \frac{{a}_{i} y_{i, \myt}}{g(\boldsymbol{x}_{i, \myt}^{\T} \beta_\myt )} - \frac{{a}_{i} y_{i, \myt}}{g(\boldsymbol{x}_{i, \myt}^{\T} \hat \beta_\myt^{\ \ell_1\rm{TCL}} )} \right| \leq \frac{1}{n_\myt} \sum_{i = 1}^{n_\myt} \frac{ M_Y |g(\boldsymbol{x}_{i, \myt}^{\T} \beta_\myt ) - g(\boldsymbol{x}_{i, \myt}^{\T} \hat \beta_\myt^{\ \ell_1\rm{TCL}} )|}{m_g\left(m_g - |g(\boldsymbol{x}_{i, \myt}^{\T} \beta_\myt ) - g(\boldsymbol{x}_{i, \myt}^{\T} \hat \beta_\myt^{\ \ell_1\rm{TCL}} )|\right)}.
    \end{split}
\end{equation}
Since $g(\cdot)$ is $L$-Lipschitz, we have
\begin{equation*}
    |g(\boldsymbol{x}_{i, \myt}^{\T} \beta_\myt ) - g(\boldsymbol{x}_{i, \myt}^{\T} \hat \beta_\myt^{\ \ell_1\rm{TCL}} )| \leq L |\boldsymbol{x}_{i, \myt}^{\T} (\beta_\myt -  \hat \beta_\myt^{\ \ell_1\rm{TCL}})| \leq L \norm{\boldsymbol{x}_{i, \myt}}_\infty \left\|\hat \beta_\myt^{\ \ell_1\rm{TCL}} -\beta_\myt \right\|_1,
\end{equation*}
where the last inequality comes from Hölder's inequality. Due to Assumption~\ref{A4} and the fact that $f(x) = {x}/{(m_g - x)}$ monotonically increase on a domain $0 \leq x < m_g$, we can further bound the right-hand side (RHS) of \eqref{eq:pf3} as follows:
\begin{equation}
    \begin{split}\label{eq:pf4}
        \left|\frac{1}{n_\myt} \sum_{i = 1}^{n_\myt} \frac{{a}_{i} y_{i, \myt}}{g(\boldsymbol{x}_{i, \myt}^{\T} \beta_\myt )} - \frac{{a}_{i} y_{i, \myt}}{g(\boldsymbol{x}_{i, \myt}^{\T} \hat \beta_\myt^{\ \ell_1\rm{TCL}} )} \right| & \leq \frac{1}{n_\myt} \sum_{i = 1}^{n_\myt} \frac{M_X M_Y L\left\|\hat \beta_\myt^{\ \ell_1\rm{TCL}} -\beta_\myt \right\|_1}{m_g\left(m_g - M_X L\left\|\hat \beta_\myt^{\ \ell_1\rm{TCL}} -\beta_\myt \right\|_1\right)} \\
        & \leq \frac{1}{n_\myt} \sum_{i = 1}^{n_\myt} \frac{M_X M_Y L\left\|\hat \beta_\myt^{\ \ell_1\rm{TCL}} -\beta_\myt \right\|_1}{m_g^2/2}.
    \end{split}
\end{equation}
The above inequality will hold since, for large enough $n_\myt, n_\mys$ and in the regime \eqref{eq:regime}, Lemma~\ref{lma:TL-beta} guarantees $$M_X L\left\|\hat \beta_\myt^{\ \ell_1\rm{TCL}} -\beta_\myt \right\|_1 \rightarrow 0,$$ and therefore we will have $M_X L\left\|\hat \beta_\myt^{\ \ell_1\rm{TCL}} -\beta_\myt \right\|_1 \leq m_g/2$.
Similarly, we can obtain
\begin{equation}
    \begin{split}\label{eq:pf5}
        \left|\frac{1}{n_\myt} \sum_{i = 1}^{n_\myt} \frac{(1-{a}_{i}) y_{i, \myt}}{1-g(\boldsymbol{x}_{i, \myt}^{\T} \beta_\myt )} - \frac{(1-{a}_{i}) y_{i, \myt}}{1 - g(\boldsymbol{x}_{i, \myt}^{\T} \hat \beta_\myt )} \right| \leq \frac{1}{n_\myt} \sum_{i = 1}^{n_\myt} \frac{M_X M_Y L\left\|\hat \beta_\myt^{\ \ell_1\rm{TCL}} -\beta_\myt \right\|_1}{m_g^2/2}.
    \end{split}
\end{equation}

Now, \eqref{eq:pf1} and \eqref{eq:pf2} tell us that, with probability as least $1 - 1/n_{\myt}$,
\begin{equation*}
    \begin{split}
         \left| \hat \tau_\myt^{\ \ell_1\rm{TCL}} - \tau \right|
         \leq & \left| \hat \tau_\myt^{\ \ell_1\rm{TCL}} - \frac{1}{n_\myt} \sum_{i = 1}^{n_\myt} \frac{{a}_{i} y_{i, \myt}}{g(\boldsymbol{x}_{i, \myt}^{\T} \beta_\myt )} - \frac{(1-{a}_{i}) y_{i, \myt}}{1-g(\boldsymbol{x}_{i, \myt}^{\T} \beta_\myt )} \right| \\
         & +\left| \frac{1}{n_\myt} \sum_{i = 1}^{n_\myt} \frac{{a}_{i} y_{i, \myt}}{g(\boldsymbol{x}_{i, \myt}^{\T} \beta_\myt )} - \frac{(1-{a}_{i}) y_{i, \myt}}{1-g(\boldsymbol{x}_{i, \myt}^{\T} \beta_\myt )} - \tau \right| \\
         <& \frac{20 M_X M_Y L \lambda}{m_g^2 \gamma}\left(\frac{1}{8\psi^2} + \frac{1}{\psi} + \frac{s}{\phi^2}\right) + \frac{2 M_Y}{m_g} \sqrt{\frac{\log(12n_\myt)}{n_\myt}} \\
         \leq & \frac{20 M_X M_Y L \lambda}{m_g^2 \gamma}\left(\frac{1}{8\psi^2} + \frac{1}{\psi} + \frac{1}{\phi^2}\right) s + \frac{2 M_Y}{m_g} \sqrt{\frac{\log(12n_\myt)}{n_\myt}}.
    \end{split}
\end{equation*}
Here, the last inequity above assumes $s \geq 1$, however, this inequality is merely to make the result look more concise, and we can handle the $s=0$ case in a similar manner.
Plugging the $\lambda$ choice \eqref{eq:lambda_1} into the above equation, and notice that, for any constant $\delta > 0$, for large enough $n_\myt$ the following holds:
\[ \sqrt{\log (12n_\myt)} = \sqrt{\log 12 + \log n_\myt} \leq (1 + \delta) \sqrt{\log n_\myt}, \quad \sqrt{\log(6n_\myt d)} \leq (1 + \delta) \sqrt{\log(n_\myt d)}.\]
We can obtain the non-asymptotic result in eq.~\eqref{eq:main}.
Now we complete the proof.

\end{proof}

\section{Additional Details for the Real Data Example}\label{appendix:real_exp}


\subsection{Dataset description}\label{appendix:realdata}
Our data is unique in three ways. Firstly, utility companies and system operators may collect fine-grained data on failure and restoration mainly for reporting purposes, but they are generally not shared beyond service territories \citep{Bryan2012,Campbell2012}, our data has the information of all service territories aggregated.
Secondly, most data are often aggregated into daily statistics over the entire region, too crude to study the resilience of the infrastructure and services \citep{Executive2013}, however, our data has the outage record per quarter hour for each city (351 in total) in Massachusetts. 
Lastly, those three consecutive Nor'easters in March 2018, Massachusetts left outage data under normal weather conditions in between, enabling us to study the restoration time under both severe and normal conditions.

\vspace{0.03in}

\noindent
\textbf{SAIDI.} \
Here, we do little modifications to represent the power reliability during the month (quantified in minutes) and calculate SAIDI for each city. Consider one city with $M$ power grid users, it went through $N$ outage events in the month with each outage lasting for $L_n$ hours, $n=1, \dots, N$. The number of power grid users recorded to be without electricity is $C_{ln}$ during hour $l$ for outage event $n$. Then SAIDI for the city in the month can be calculated as (separately calculated under severe/normal weather conditions):
\begin{equation}\label{eq:saidi}
    {\rm SAIDI} = \frac{\sum_{n=1}^{N}\sum_{l=1}^{L_n}C_{ln}}{M}\times 60
\end{equation}
We consider all the power outage events with a minimum outage rate bigger than 0.1\% and last for over 2 hours.

\vspace{0.03in}

\noindent
\textbf{Weather.} \
Extreme weather and climate events have emerged as primary catalysts for infrastructure damage, leading to widespread power outages and supply inadequacy risks in the United States \citep{MUKHERJEE2018283,MUKSA}. Consequently, analyzing the patterns of these outages can be facilitated by leveraging weather data.

To perform the analysis, we obtained weather data from the High-Resolution Rapid Refresh (HRRR) model \citep{HRRR}. Developed by the National Centers for Environmental Prediction (NCEP), the HRRR model is a numerical weather prediction model that provides high-resolution and frequently updated forecasts for regional weather conditions. By assimilating data from various sources, such as satellites, radars, and weather stations, the HRRR model combines observational information with advanced mathematical equations to simulate atmospheric behavior. This enables the model to generate highly detailed and accurate short-term forecasts, ranging from 1 to 18 hours, with a spatial resolution as fine as 3 kilometers. Hourly weather data was collected from all available stations, and for each city, we matched the data with the nearest stations to obtain the corresponding weather information. In our analysis, we specifically focused on two key weather parameters extracted from the HRRR model: wind speed (m/s) and precipitable water (kg/$m^2$). 

In the event of a power outage, the categorization of weather conditions as 'severe' is determined based on specific thresholds: a maximum wind speed exceeding 19 m/s, or a maximum precipitable water value surpassing 16 kg/$m^2$. If neither of these criteria is met, the weather condition is classified as 'normal'. This dichotomous classification facilitates the assessment of the SAIDI by distinguishing between outages that happened under varying weather conditions.

\vspace{0.03in}

\noindent
\textbf{Demographic factors.} \
Previous research in the field of power equity has extensively examined the correlation between demographic factors and power outages. Studies have demonstrated that certain disadvantaged communities, such as low-income and minority communities, experience a disproportionate impact from power outages due to their limited resources for recovery \citep{LIN2022107135,UrbanR}.

To investigate the demographic factors influencing power outages, we leverage the comprehensive data provided by the American Community Survey (ACS) 5-Year Data. This dataset is a continuous survey conducted by the U.S. Census Bureau, focusing on gathering in-depth information about the demographic, social, economic, and housing characteristics of the U.S. population. The ACS 5-Year Data is particularly valuable as it encompasses information collected over a 5-year period, providing a more robust and accurate representation of the population and its diverse characteristics. To complement the demographic factors, we also obtained the number of power grid users from the local utilities. All the demographic factors taken into account are detailed in Table~\ref{table:demographic factors description}.
 
Demographic data is initially gathered at the census tract level. To ensure compatibility with city-level outage records, we adjust the geographical scope of these demographic factors. This is achieved by computing the average values of each demographic factor across all census tracts that either fall within or intersect with the boundaries of the specific city.

\begin{table}[!htp]
\centering
    \caption{\small Demographic factors and the descriptions}
    \label{table:demographic factors description}
\begin{tabular}{@{}ll@{}}
\toprule
\textbf{Attribute Name} & \textbf{Description}                                                  \\ \midrule
Median Income          & Median household Income (dollars)                                     \\
TotalCustomer          & Number of power grid users in the city                                \\
ElderlyPercentage       & percentage of 60 years and over individuals                          \\
ParticipationRate      & Labor force participation rate                                        \\
UnemploymentRate       & Unemployment rate of individuals over 16                               \\
WhitePercentage        & Percentage of the white people                                        \\
MinorityPercentage      & Percentage of the black and African American                          \\
AsianPercentage        & Percentage of the Asian                                               \\
MedianHousingValue     & Median Value of Owner-occupied housing units                          \\
EducationalAttainment  & Percentage of individuals over 18 with Bachelor's degree or higher    \\
PovertyPercentage & Percentage below the poverty level (among the population \\
& for whom poverty status is determined) for the past 12 months \\
MalePercentage         & Percentage of the male                                                \\
MedianAge              & Median age of the individuals                                         \\
OldAgeDependencyRatio  & Percentage of the elderly population to the working-age population    \\
ACPercentage           & Percentage of agriculture, forestry, fishing, hunting, and mining     \\
USCitizenPercentage    & Percentage of US citizen                                              \\
ForeignBornPercentage  & Percentage of the foreign borns                                       \\
DisabledPercentage     & Disability percentage of the civilian noninstitutionalized population \\
SchoolEnrollment       & Percentage of individuals over 3 enrolled in school                   \\
MedianRent             & Median contract rent                                                  \\
RentPercentage         & Median gross rent as a percentage of household income                 \\
InsuredPercentage      & Percentage of the insured of civilian noninstitutionalized population \\
 \bottomrule
\end{tabular}
\end{table}

\vspace{0.03in}

\noindent
\textbf{Land cover and land use.} \
Several studies \citep{ctx8127054220004436,LCQ} have indicated that land cover and land use variables can serve as reasonable proxies for power systems data, such as the number of poles. This suggests that land cover variables have the potential to serve as a valuable tool in developing generalized outage models that can be applied to service areas lacking detailed power system data. For our analysis, we obtained land cover and land use data from the Massachusetts government's MassGIS Data: 2016 Land Cover/Land Use (LCLU) dataset \citep{LCLU}. This statewide dataset combines land cover mapping from 2016 aerial imagery with land use information derived from standardized assessor parcel data specific to Massachusetts. The creation of this dataset was a collaborative effort between MassGIS and the National Oceanic and Atmospheric Administration's (NOAA) Office of Coastal Management (OCM).

In our study, we considered various land cover categories, including forest, grass, and wetland. Additionally, we examined different land use types, such as single-family residential, multi-family residential, and agriculture. Each of these land cover and land use variables was measured as a percentage of area for each city in the dataset.

\subsection{Visualization of the outage records on the Massachusetts map} 
\begin{figure}[!htb]
\centering
      \includegraphics[width = .85\textwidth]{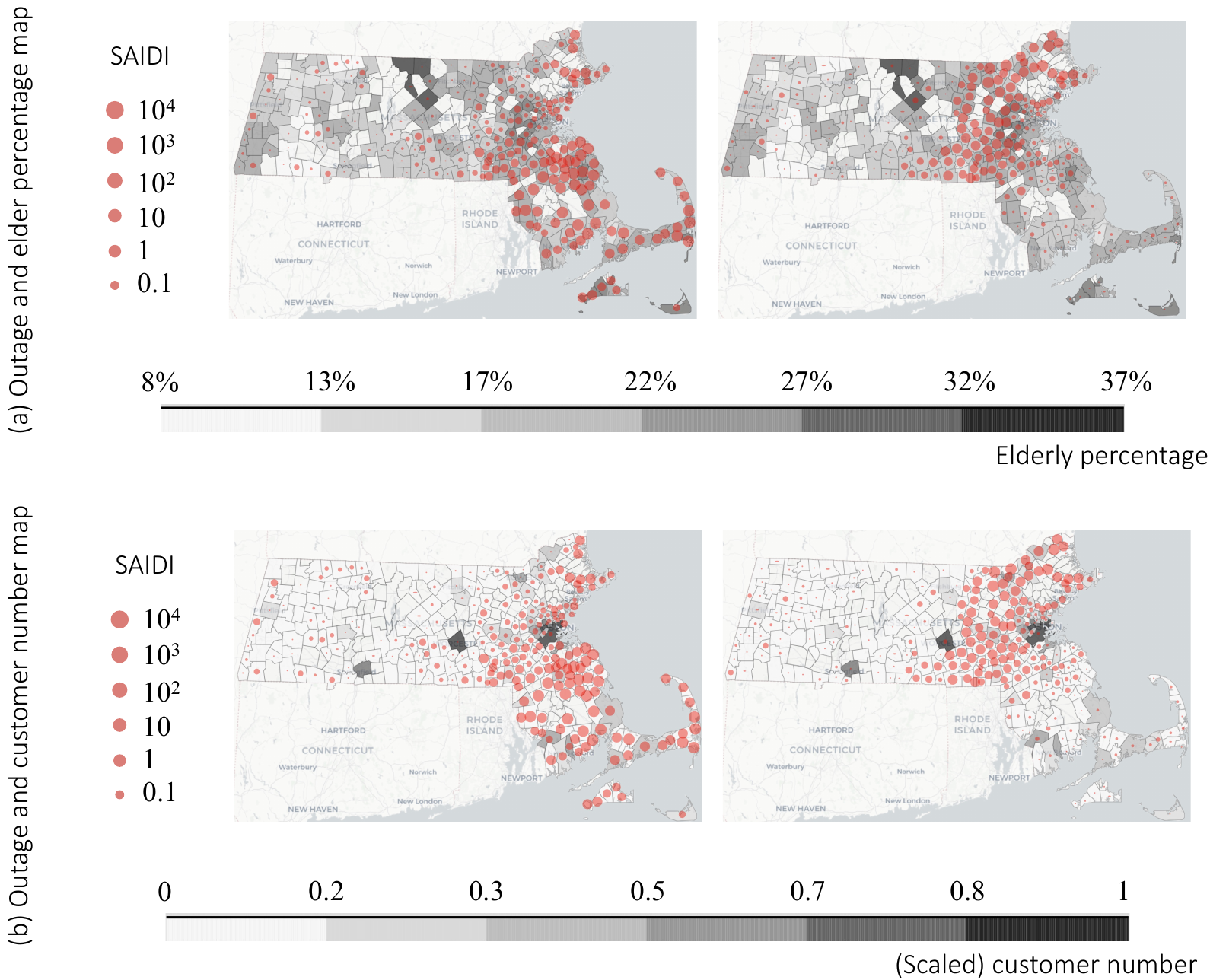}
  \caption{\small Data visualization on the Massachusetts map: The background color gradient, ranging from light to dark grey, illustrates the magnitude of the protected attributes; The size of the red bubbles corresponds to the magnitude of SAIDI, with larger bubbles indicating higher SAIDI values. Sub-figure (b) presents a standardized customer number, scaled between 0 and 1, to maintain data sensitivity, contrasting with the real data range shown in sub-figure (a).}\label{fig:map}
\end{figure}

In addition to Figure~\ref{fig:counterfactual_illus}(b), we visualize the other raw protected attributes and the outcome variable, \ie, the SAIDI, on the Massachusetts map in Figure~\ref{fig:map}.
As shown in Figure~\ref{fig:map}(a), cities with the highest percentages of elderly populations endure lower SAIDI, with lower SAIDI also observed in cities with lower elderly percentages. This observation suggests a complex relationship between the elderly population percentage and SAIDI that mere correlation analysis might not reveal. According to Figure~\ref{fig:map}(b), larger SAIDI values are more common in cities with fewer power grid customers, under both weather conditions. The distribution of higher SAIDI varies with weather, being more prevalent on the southeast coast in severe weather but further west inland in normal conditions. Our $\ell_1$-TCL method, however, indicates that this variation in SAIDI relative to user density is primarily evident in severe weather conditions.

\subsection{Additional results for hyperparameter selection}\label{sec:hyp_selection}

\begin{figure}[!htp]
\centerline{\includegraphics[width = \textwidth]{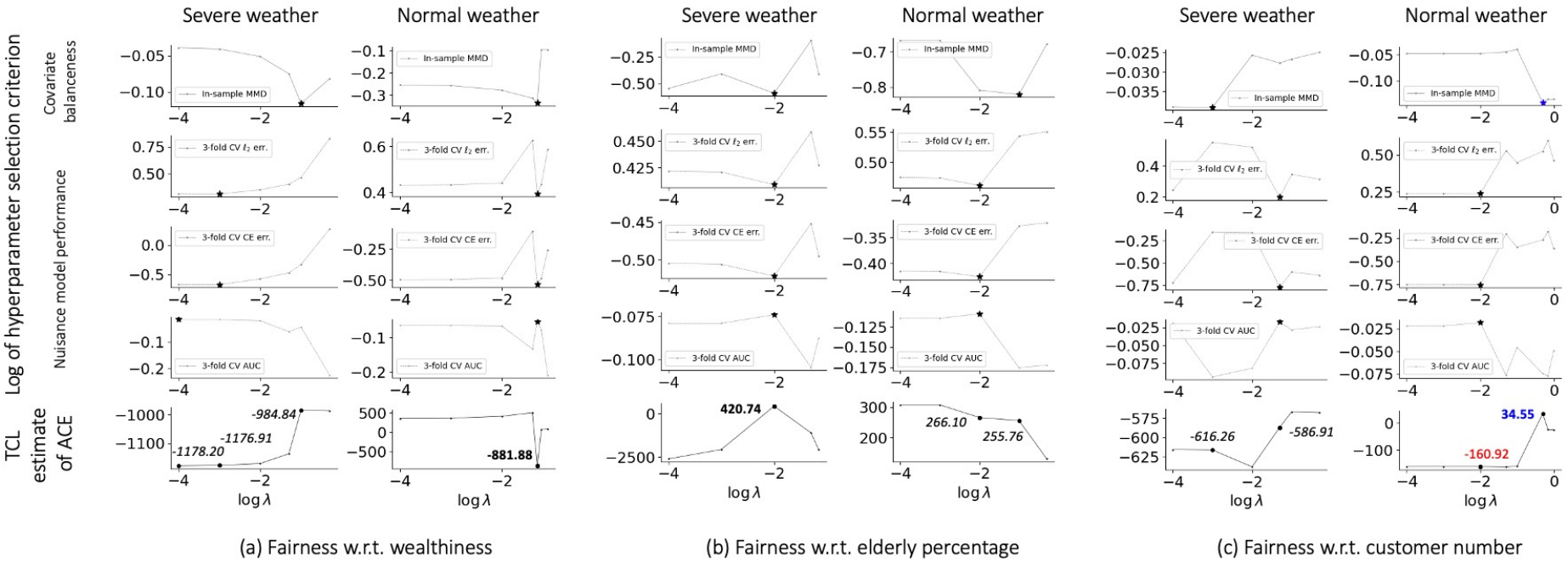} 
}
\caption{\small Hyperparameter selection for unfairness assessment w.r.t. wealthiness (left), elderly percentage (middle), and customer number (right): The first 4 rows report the selection criterion (we use ``$\star$'' to mark the selected hyperparameter) and the last row reports the resulting $\ell_1$-TCL estimates of the ACE, where we use ``$\bullet$'' to denote the corresponding $\ell_1$-TCL estimate in the last row (results using MMD criterion is in bold font). We use \emph{italic} font to highlight the selected $\ell_1$-TCL estimates using different criteria, showing that these criteria output very similar results; \textbf{Bold} font indicates that different criteria output the same $\ell_1$-TCL estimate; In the customer number (right) cases, we use different colors for $\ell_1$-TCL estimate with MMD criterion ({\color{blue} blue}) and nuisance model performance criterion ({\color{red} red}) to highlight that our proposed MMD criterion outputs results different from existing nuisance model performance criteria.}
\label{fig:realExpMAHypSelection}
\end{figure}

For completeness, we report how the selection criteria and the estimated ACE vary with different $\ell_1$ regularization strength $\lambda$'s in Figure~\ref{fig:realExpMAHypSelection}.

It is worthwhile noting that the similar performance of different selection criteria is again observed when we consider the elderly percentage as the protected attribute, reaffirming their effectiveness and supporting the enhanced reliability and accuracy of the $\ell_1$-TCL results. 
Meanwhile, when we take the customer number as the protected attribute, we can observe that the in-sample MMD criterion can output very different results (highlighted in blue) compared with CV nuisance model performance. On one hand, the numerical simulation shows that in-sample MMD criterion can help yield a more accurate causal effect estimate; On the other hand, it agrees with common sense that power company will prioritize the power supply in populated areas under sever conditions, which is explainable discrimination and does not contribute to unfairness, and such a power supply difference does not exist under normal conditions (which, again, is not captured by the vanilla IPW estimator).


\subsection{Additional results for uncertainty quantification}\label{appendix:BPmore}
We only perform around 15 repetitions since each bootstrap sample requires its unique hyperparameter grid: Even though we take grid length to be $10^{-3}$, resulting in over $10^3$ $\lambda$'s to select from, we still need to manually adjust the exact grid positions for each Bootstrap trial to output reasonable results.

\begin{figure}[!htp]
\centerline{\includegraphics[width = \textwidth]{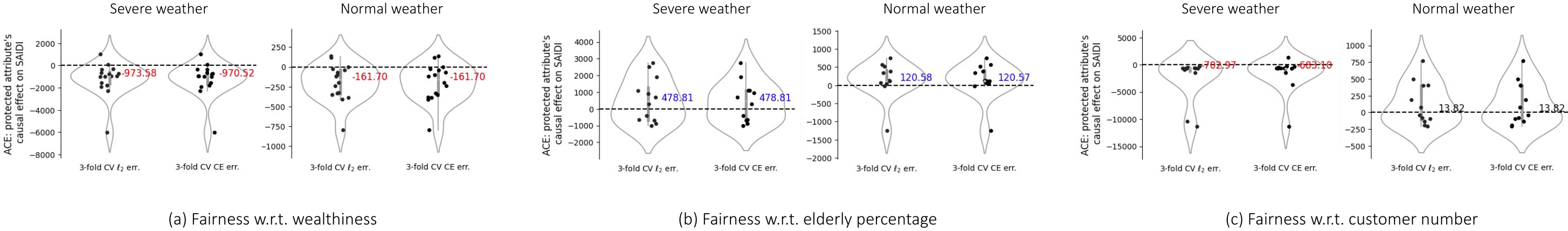}}
\caption{\small Bootstrap UQ visualization of our $\ell_1$-TCL incorporating additional selection criteria for unfairness assessment w.r.t. protected attributes (as detailed in the sub-captions on the bottom).  Again, {\color{blue} blue} (median ACE $>-100$) represents a positive causal effect, {\color{red} red} (median ACE $<-100$) indicates a negative effect, and black ($-100 \leq$ median ACE $\leq 100$) suggests a neutral effect. These results agree with the $\ell_1$-TCL results presented in Figure~\ref{fig:realExpMABP}, re-affirming the reliability of our $\ell_1$-TCL.}
\label{fig:realExpMABPmore}
\end{figure}

For completeness, we report UQ results using other nuisance model performance criteria in Figure~\ref{fig:realExpMABPmore}, where we observe similar pattern shown in Figure~\ref{fig:realExpMABP}: On one hand, our $\ell_1$-TCL consistently generates reasonable findings that agrees with existing literature. On the other hand, $\ell_1$-TCL using different selection criteria can yield very similar results, which supports the reliability and robustness of our proposed $\ell_1$-TCL.



\end{document}